\documentclass[10pt,twocolumn,letterpaper]{article}

\usepackage{iccv}
\usepackage{times}
\usepackage{epsfig}
\usepackage{graphicx}
\usepackage{amsmath}
\usepackage{amssymb}
\usepackage{floatrow}
\usepackage{adjustbox}
\usepackage[utf8]{inputenc} %
\usepackage[T1]{fontenc}    %
\usepackage{url}            %
\usepackage{booktabs}       %
\usepackage{amsfonts}       %
\usepackage{nicefrac}       %
\usepackage{microtype}      %
\usepackage{mwe}
\usepackage{makecell}
\usepackage[accsupp]{axessibility}
\usepackage{wrapfig}
\usepackage{algorithm}
\usepackage{stfloats}
\usepackage[font=small,skip=0pt]{caption}
\usepackage[noend]{algpseudocode}
\usepackage{paralist}
\usepackage{multirow}
\usepackage{color, colortbl}
\usepackage{xcolor}
\usepackage{xspace}
\usepackage{etoc}
\usepackage{subcaption}
\usepackage{mwe}
\usepackage{comment}
\usepackage[numbers,sort,compress]{natbib}

\definecolor{Gray}{gray}{0.9}

\DeclareCaptionLabelFormat{andtable}{\tablename~\thetable}

\newfloatcommand{capbtabbox}{table}
\newfloatcommand{capbfigbox}{figure}
\usepackage[breaklinks=true,bookmarks=false,colorlinks]{hyperref}

\newcommand{\method}{\texttt{CLUE}\xspace}
\newcommand{\source}{\mathcal{S}}  %
\newcommand{\target}{\mathcal{T}}  %
\newcommand{\budget}{B}  %

\newcommand{\numclasses}{C}  %
\newcommand{\featext}{\phi}  %
\newcommand{\model}{h}  %
\newcommand{\modelparams}{\Theta}  %
\newcommand{\setpartition}{\mathcal{S}}  %
\newcommand{\numclusters}{K}
\newcommand{\cluster}{k}
\newcommand{\timestep}{\rho}
\newcommand{\unlabeled}{\mathcal{U}}
\DeclareMathOperator*{\argmin}{argmin}

\iccvfinalcopy %

\def\iccvPaperID{1712} %

\begin{document}

\title{Active Domain Adaptation via Clustering Uncertainty-weighted Embeddings}

\author{
    \textbf{Viraj Prabhu}$^{1}$ \qquad
    \textbf{Arjun Chandrasekaran}\thanks{Work done partially at Georgia Tech.}$^{\,\,,2}$ \qquad
    \textbf{Kate Saenko}$^3$ \qquad 
    \textbf{Judy Hoffman}$^{1}$ \qquad \\
    $^1$Georgia Tech \qquad
    $^2$Max Planck Institute for Intelligent Systems, Tübingen \qquad
    $^3$Boston University \\
    {\small\texttt{\{virajp,judy\}@gatech.edu} \qquad \texttt{ achandrasekaran@tue.mpg.de} \qquad \texttt{saenko@bu.edu}}
}

\maketitle

\begin{abstract}
   \vspace{-10pt}
   Generalizing deep neural networks to new target domains is critical to their real-world utility. In practice, it may be feasible to get some target data labeled, but to be cost-effective it is desirable to select a maximally-informative subset via active learning (AL). We study the problem of AL under a domain shift, called Active Domain Adaptation (Active DA). We demonstrate how existing AL approaches based solely on model uncertainty or diversity sampling are less effective for Active DA. We propose Clustering Uncertainty-weighted Embeddings (\method), a novel label acquisition strategy for Active DA that performs uncertainty-weighted clustering to identify target instances for labeling that are both uncertain under the model and diverse in feature space. %
   \method consistently outperforms competing label acquisition strategies for Active DA and AL across learning settings on 6 diverse domain shifts for image classification. Our code is available at \url{https://github.com/virajprabhu/CLUE}.
   \vspace{-7pt}
\end{abstract}

\vspace{-12pt}
\section{Introduction}
\label{sec:intro}
\vspace{-5pt}

Deep neural networks excel at learning from large labeled datasets but struggle to generalize to 
new target domains~\cite{saenko2010adapting,torralba2011unbiased}. This limits their real-world utility, as it is impractical to collect a large new dataset for every new deployment domain. Further, all target instances are usually \emph{not} equally informative, and it is far more cost-effective to identify maximally informative target instances for labeling. While Active Learning~\cite{ash2019deep,cohn1994improving,ducoffe2018adversarial,gal2017deep,sener2017active,settles2009active} has extensively studied the problem of identifying informative instances for labeling, it typically focuses on learning a model from scratch and does not operate under a domain shift. In many practical scenarios, models are trained in a source domain and deployed in a different target domain, often with additional domain adaptation~\cite{ganin2014unsupervised,hoffman2017cycada,saenko2010adapting,tzeng2014deep}. In this work, we study the problem of active learning under such a domain shift, called Active Domain Adaptation~\cite{rai2010domain} (Active DA).

Concretely, given i) labeled data in a source domain, ii) unlabeled data in a target domain, and iii) the ability to obtain labels for a fixed budget of target instances, the goal of Active DA is to select target instances for labeling and learn a model with high accuracy on the target test set. Active DA has widespread utility as a means of cost-effective adaptation from cheaper to more expensive sources of labels (\eg synthetic to real data), as well as when the quantity (\eg autonomous driving) or cost (\eg medical diagnosis) of labeling in the target domain is prohibitive. Despite its practical utility, it is a challenging task that has seen limited follow-up work since its introduction over ten years ago~\cite{chattopadhyay2013joint,rai2010domain,su2019active}.

\begin{figure}[t]    
    \centering
    \includegraphics[width=0.9\linewidth]{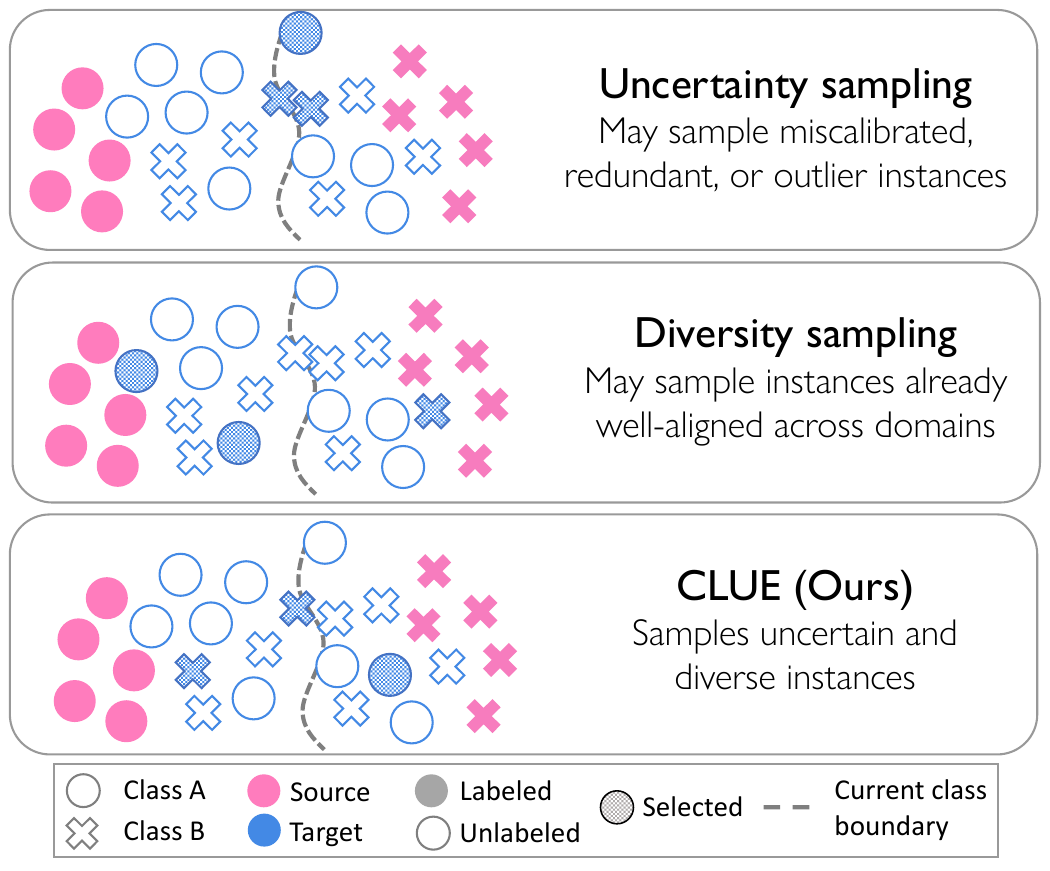}
    \caption{
    The goal of \textbf{Active Domain Adaptation}~\cite{rai2010domain} (Active DA) is to adapt a source model to an unlabeled target domain by acquiring labels for selected target instances via an oracle. Existing active learning (AL) methods based solely on uncertainty~\cite{gal2017deep,roth2006margin,wang2014new} or diversity-sampling~\cite{gissin2019discriminative,sener2017active} are less effective for Active DA (Row 1, 2). We propose \method, an AL method designed for Active DA that selects instances that are both uncertain (thus informative to the model) and diverse in feature space (thus minimizing redundancy, Row 3), and leads to more cost-effective adaptation than competing AL and Active DA methods (Sec.~\ref{subsec:results}).}
    \label{fig:teaser}    
    \vspace*{-7pt}
\end{figure}

\vspace{-2pt}
The traditional AL setting typically focuses on techniques to select samples to efficiently learn a model from scratch, rather than adapting under a domain shift~\cite{settles2009active}. As a result, existing state-of-the art AL methods based on either uncertainty or diversity sampling are less effective for Active DA. Uncertainty sampling selects instances that are highly uncertain under the model's beliefs~\cite{gal2017deep,ducoffe2018adversarial,tong2001support,kirsch2019batchbald}.
Under a domain shift, uncertainty estimates on the target domain may be miscalibrated~\cite{snoek2019can} and lead to sampling uninformative, outlier, or redundant instances (Fig.~\ref{fig:teaser}, top). A parallel line of work in AL based on diversity sampling instead selects instances dissimilar to one another in a learned embedding space~\cite{gissin2019discriminative,sener2017active,sinha2019variational}. In Active DA, this can lead to sampling uninformative instances from regions of the feature space that are already well-aligned across domains (Fig.~\ref{fig:teaser}, middle). As a result, solely using uncertainty or diversity sampling is suboptimal for Active DA, as we demonstrate in Sec~\ref{subsec:results}.

\vspace{-2pt}
Recent work in AL and Active DA has sought to combine uncertainty and diversity sampling. \texttt{AADA}~\cite{su2019active}, the state-of-the-art Active DA method, combines uncertainty with diversity measured by `targetness' under a learned domain discriminator. However, targetness does not ensure that the selected instances are representative of the entire target data distribution (\ie not outliers), or dissimilar to one another. 
Ash~\etal~\cite{ash2019deep} instead propose performing clustering in a hallucinated ``gradient embedding'' space. However, they rely on distance-based clustering in high-dimensional spaces, which often leads to suboptimal results.

In this work, we propose a novel label acquisition strategy for Active DA that combines uncertainty and diversity sampling in a principled manner without the need for complex gradient or domain discriminator-based diversity measures. Our approach, Clustering Uncertainty-weighted Embeddings (\method), identifies informative and representative target instances from dense regions of the feature space. To do so, \method clusters deep embeddings of target instances \emph{weighted} by the corresponding uncertainty of the target model. Our weighting scheme effectively increases the density of instances proportional to their uncertainty. To construct non-redundant batches, \method then selects nearest neighbors to the inferred cluster centroids for labeling. Our algorithm then leverages the acquired target labels and, optionally, the labeled source and unlabeled target data, to update the model, consistently leading to more cost-effective domain alignment than competing (and frequently more complex) alternatives. 

\vspace{-2pt}
\noindent \textbf{Contributions:}
\begin{compactenum}
\item We benchmark the performance of state-of-the-art methods for active learning on challenging domain shifts, and find that methods based purely on uncertainty or diversity sampling are not effective for Active DA.
\item We present \method, a novel and easy-to-implement label acquisition strategy for Active DA that uses uncertainty-weighted clustering to identify instances that are both uncertain under the model and diverse in feature space. 
\item We present results on 6 diverse domain shifts from the DomainNet~\cite{peng2019moment}, Office~\cite{saenko2010adapting}, and DIGITS ~\cite{netzer2011reading,lecun1998gradient} benchmarks for image classification.
Our method \method improves upon both the previous state-of-the art in Active DA across shifts (by as much as 9\% in some cases), as well as state-of-the-art methods for active learning, across multiple learning strategies.
\end{compactenum}

\vspace{-3pt}
\section{Related Work}
\label{sec:relwork}
\vspace{-5pt}

\begin{figure*}[t]
    \centering
    \includegraphics[width=\textwidth]{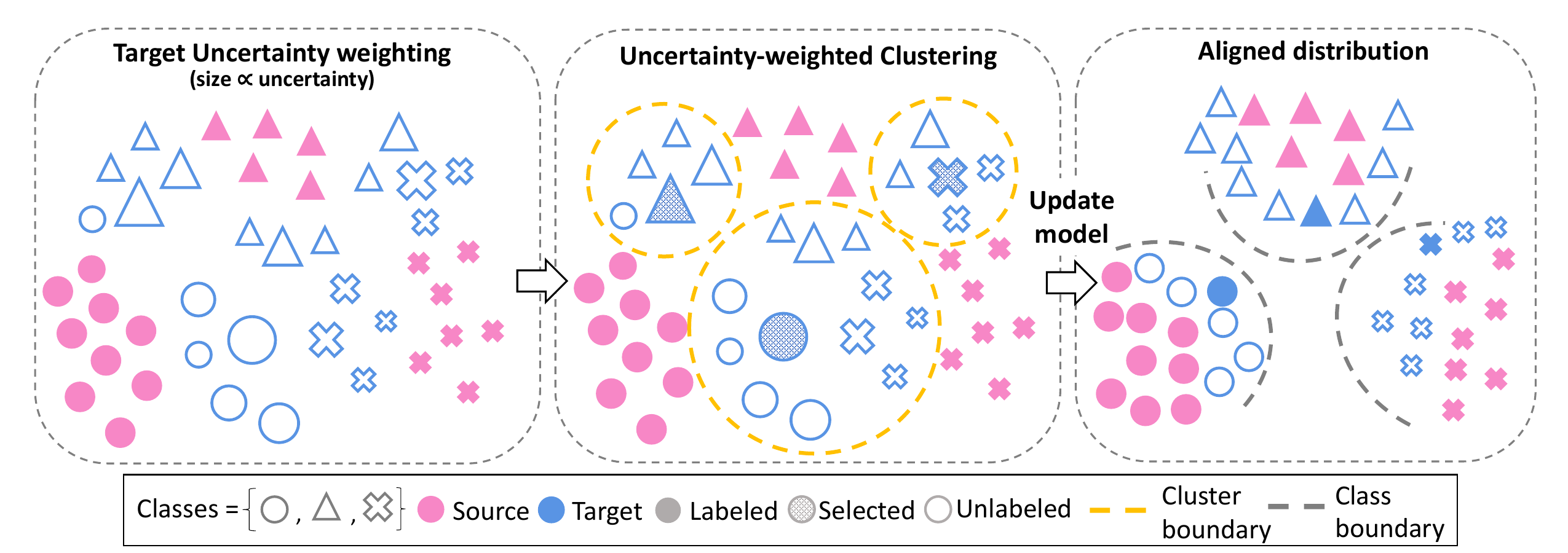}
    \caption{
    We propose CLustering Uncertainty-weighted Embeddings ($\method$), a novel label acquisition strategy for Active DA that identifies a diverse set of target instances that are informative and representative (Eq.~\ref{eq:clue}). First, deep embeddings of target instances are reweighted based on model entropy to emphasize uncertain regions of feature space (\emph{left}). Next, to select diverse instances, these uncertainty-weighted embeddings are clustered, and the instance closest to each cluster centroid is acquired for labeling (\emph{middle}). Finally, the acquired target labels (and optionally, the labeled source and unlabeled target data) are used to update the model, leading to well-classified target data (\emph{right}).
    }
    \label{fig:approach}
    \vspace{-5pt}
  \end{figure*}

\noindent\textbf{Active Learning (AL) for CNN's}. AL for CNN's has focused on the batch-mode setting due to the instability associated with single-instance updates. The two most successful paradigms in AL have been uncertainty sampling and diversity sampling~\cite{ash2019deep}. Uncertainty-based methods select instances with the highest uncertainty under the current model~\cite{gal2017deep,ducoffe2018adversarial,tong2001support,schohn2000less}, using measures such as entropy~\cite{wang2014new}, classification margins~\cite{roth2006margin}, or confidence. Diversity-based methods select instances that are representative of the entire dataset, and optimize for diversity in a learned embedding space, via clustering, or core-set selection~\cite{sener2017active,gissin2019discriminative,sinha2019variational,geifman2017deep}. 

Some approaches combine these two paradigms~\cite{ash2019deep,baram2004online,hsu2015active,zhdanov2019diverse}. Active Learning by Learning~\cite{hsu2015active} formulates this as a multi-armed bandit problem of selecting between coreset and uncertainty sampling at each step. Zhdanov et al.~\cite{zhdanov2019diverse} use K-Means clustering~\cite{hartigan1979algorithm} to increase batch diversity after pre-filtering based on uncertainty. More recently, BADGE~\cite{ash2019deep} runs KMeans++ on hallucinated ``gradient embeddings''. We propose \method, an AL method for sampling under a domain shift, that uses uncertainty-weighted clustering to select diverse and informative target instances.

\noindent\textbf{Domain Adaptation}. The task of transferring models trained on a labeled source domain to an unlabeled~\cite{saenko2010adapting,tzeng2014deep,ganin2014unsupervised,hoffman2017cycada} or partially-labeled~\cite{donahue2013semi,yao2015semi,saito2019semi} target domain has been studied extensively. Initial approaches aligned feature spaces by optimizing discrepancy statistics between the source and target~\cite{tzeng2014deep,long2015learning}, while in recent years adversarial learning of a feature space encoder alongside a domain discriminator has become a popular alignment strategy~\cite{ganin2014unsupervised,ganin2016domain,tzeng2017adversarial}. %
In this work, we propose a label acquisition strategy for active learning under a domain shift that generalizes across multiple domain adaptation strategies.

\noindent\textbf{Active Domain Adaptation (Active DA)}. Unlike semi-supervised domain adaptation which assumes labels for a \emph{random} subset of target instances, Active DA focuses on \emph{selecting} target instances to label for domain adaptation. Rai et al.~\cite{rai2010domain} first studied the task of Active DA applied to sentiment classification from text data. They propose ALDA, which samples instances based on model uncertainty and a learned domain separator. Chattopadhyay et al.~\cite{chattopadhyay2013joint} select target instances and learn importance weights for source points by solving a convex optimization problem of minimizing MMD between features. More recently, Su et al.~\cite{su2019active} study Active DA in the context of deep CNN's and propose AADA, wherein target instances are selected based on predictive entropy and targetness measured by an adversarially trained domain discriminator, followed by adversarial domain adaptation via \texttt{DANN}~\cite{ganin2016domain}. We propose \method, a novel label acquisition strategy for Active DA that identifies uncertain and diverse instances for labeling that outperforms prior work on diverse shifts across multiple learning strategies.
  
\vspace{-4pt}
\section{Approach}
\label{sec:approach}
\vspace{-3pt}

We address active domain adaptation (Active DA), where the goal is to generalize a model trained on a source domain to an unlabeled target domain, with the option to query an oracle for labels for a subset of target instances. While individual facets of this task -- adapting to a new domain and selective acquisition of labels, have been well-studied as the problems of Domain Adaptation (DA) and Active Learning (AL) respectively, Active DA presents the new challenge of identifying target instances that will, once labeled, result in the most \emph{sample-efficient domain alignment}.
Further, the answer to this question may vary based on the properties of the specific domain shift. %
In this section, %
we present \method, a novel label acquisition strategy for Active DA which performs consistently well across diverse domain shifts.%

\subsection{Notation \& Preliminaries}

In Active DA, the learning algorithm has access to labeled instances from the source domain $(X_\source, Y_\source)$ (solid pink in Fig.~\ref{fig:approach}), unlabeled instances from the target domain $X_{\unlabeled\target}$ (blue outline in Fig.~\ref{fig:approach}), and a budget $\budget$ ($ = 3$ in Fig.~\ref{fig:approach}) which is much smaller than the amount of unlabeled target data. 
The learning algorithm may query an oracle to obtain labels for at most $\budget$ instances from $X_{\unlabeled\target}$, and add them to the set of labeled target instances $X_{\mathcal{L}\target}$. The entire target domain data is $X_\target = X_{\mathcal{L}\target} \cup X_{\unlabeled\target}$. 
The task is to learn a function $\model: X \to Y$ (a convolutional neural network (CNN) parameterized by $\modelparams$) that achieves good predictive performance on the target.
In this work, we consider Active DA in the context of $C$-way image classification -- the samples $\mathbf{x}_\source \in X_\source$, $\mathbf{x}_\target \in X_\target$ are images, and labels $y_\source \in Y_\source$, $y_\target \in Y_\target$ are categorical variables $y \in \{1, 2, .. , \numclasses \}$. 

\noindent \textbf{Active Learning.} The goal of active learning (AL) is to identify target instances that, once labeled and used for training the model, minimize its expected future loss. In practice, prior works in AL identify such instances based primarily on two proxy measures, uncertainty and diversity (see Sec.~\ref{sec:relwork}). We first revisit these terms in the context of Active DA.

\noindent\textbf{Uncertainty.} 
Prior work in AL has proposed using several measures of model uncertainty as a proxy for informativeness (see Sec.~\ref{sec:relwork}). 
However, in the context of Active DA, using model uncertainty to select informative samples presents a conundrum. On the one hand, models benefit from initialization on a related source domain rather than learning from scratch. On the other hand, under a strong distribution shift, model uncertainty may often be miscalibrated~\cite{snoek2019can}. Unfortunately however, without access to target labels, it is impossible to evaluate the reliability of model uncertainty!

\noindent\textbf{Diversity.} 
Acquiring labels solely based on uncertainty often leads to sampling batches of similar instances with high redundancy, or to sampling outliers.
A parallel line of work in active learning instead proposes sampling \emph{diverse} instances that are representative of the unlabeled pool of data. 
Several definitions of ``diverse'' exist in the literature: some works define diversity as coverage in feature~\cite{sener2017active} or ``gradient embedding'' space~\cite{ash2019deep}, while prior work in Active DA measures diversity by how ``target-like'' an instance is~\cite{su2019active}.
In Active DA, training on a related source domain (optionally followed by unsupervised domain alignment), results in some classes being better aligned across domains than others. Thus, in order to be cost-efficient it is important to avoid sampling from already well-learned regions of the feature space. However, purely diversity-based AL methods are unable to account for this, and lead to sampling redundant instances.

While sampling instances that are either uncertain or diverse may be useful to learning, an optimal label acquisition strategy for Active DA would ideally capture both jointly. We now introduce \method, a label acquisition strategy for Active DA that captures both uncertainty and diversity.

\subsection{Clustering Uncertainty-weighted Embeddings}
\label{subsec:clue}

To measure informativeness we use predictive entropy $\mathcal{H}(Y|\mathbf{x}; \modelparams)$~\cite{wang2014new} ($\mathcal{H}(Y|\mathbf{x})$ for brevity), which for $C$-way classification, is defined as: 
\vspace{-5pt}
\begin{equation}
	\mathcal{H}(Y|\mathbf{x}) 
	=  - \sum_{c=1}^{\numclasses} p_{\modelparams}(Y=c | \mathbf{x}) \log p_{\modelparams}(Y=c | \mathbf{x})  
\label{eq:ent}
\vspace{-5pt}
\end{equation}
\vspace{-5pt}

Under a domain shift, entropy can be viewed as capturing both uncertainty \emph{and} domainness. Rather than training an explicit domain discriminator~\cite{ganin2014unsupervised,su2019active}, we consider an \emph{implicit} domain classifier $d(\mathbf{x})$~\cite{saito2019semi} based on entropy thresholding~\footnote{This assumes that source instances have low predictive entropy, which is generally satisfied under supervised training.}:
\vspace{-2pt}
\begin{equation}
  d(\mathbf{x})=\left\{\begin{array}{ll}1, & \text { if } \mathcal{H}(Y|\mathbf{x})) \geq \gamma \\ 0, & \text { otherwise }\end{array}\right.
  \label{eq:dom}  
\end{equation}
\vspace{-2pt}

\noindent where 1 and 0 denote target and source domain labels, and $\gamma$ is a threshold value. The probability of an instance belonging to the target domain is thus given by: 
\begin{equation}
\begin{aligned}
  \vspace{-5pt}
p(d(\mathbf{x})\!=\!1) & \!=\! \frac{\mathcal{H}(Y|\mathbf{x})}{\log(C)}
\!\propto\! \mathcal{H}(Y|\mathbf{x}) \;\;\; \left[C\,\text{is constant}\right]
\vspace{-3pt}
\end{aligned}
\end{equation}
where $\log(C)$ is the maximum possible entropy of a $C-$way distribution. Next, we measure diversity based on feature-space coverage. Let $\featext(\mathbf{x})$ denote feature embeddings extracted from model $\model$. We identify diverse instances by partitioning $X_T$ into $K$ diverse sets via a partition function $\setpartition: X_T \rightarrow \{X_1, X_2, ..., X_K\}$
. Let $\{\mu_1, \mu_2, ..., \mu_K\}$ denote the corresponding centroid of each set. Each set $X_k$ should have a small variance $\sigma^2(X_k)$. 
Expressed in terms of pairs of samples,
$\sigma^2(X_k) = \frac{1}{2|X_k|^2}\sum_{\mathbf{x_i}, \mathbf{x_j} \in X_k}  || \phi(\mathbf{x_i}) - \phi(\mathbf{x_j})||^2$~\cite{zhang2012some}. 
The goal is to group target instances that are similar in the CNN's feature space, into a set $X_k$. However, while $\sigma^2(X_k)$ is a function of the target data distribution and feature space $\phi(.)$, it does not account for uncertainty.

\begin{algorithm}[t]
  \begin{algorithmic}[1]
  \State {\bf Require}: Neural network $\model = f(\featext(.))$, parameterized by $\modelparams$,
              labeled source data $(X_\source, Y_\source)$, unlabeled target data $X_\target$, Per-round budget $B$, Total rounds $R$.
  \State {\bf Define:} Labeled target set $X_{\mathcal{L}\target} = \emptyset$
  \State Train source model $\modelparams^1$ on $(X_\source, Y_\source)$. %
  \State Adapt model to unlabeled target domain (optional). %
  \For{$\timestep = 1$ to $R$}
    \State \textbf{CLUE:} For all instances $\mathbf{x} \in X_\target \setminus X_{\mathcal{L}\target}$:
      \begin{enumerate}
         \item Compute deep embedding $\featext(\mathbf{x})$
         \item Run Weighted K-Means until convergence (Eq.~\ref{eq:clue}):
          \begin{enumerate}
            \item Init. $K$(=B) centroids $\{ \mathbf{\mu_{i}} \}_{i=1}^B$ (KMeans++)
            \item \textbf{Assign:} 
            \vspace{-5pt}
            \begin{equation}
              \vspace{-10pt}
              X_\cluster \gets \{ \mathbf{x} | k= \argmin_{i=1, ..., K} ||\featext(\mathbf{x}) - \mathbf{\mu_i}||^2\}_{\forall \mathbf{x}}\nonumber
            \end{equation}
            \item \textbf{Update:} $\mathbf{\mu_k} \gets \frac{\sum_{\mathbf{x} \in X_\cluster} \mathcal{H}(Y|\mathbf{x}) \featext(\mathbf{x})}{\sum_{\mathbf{x} \in X_\cluster} \mathcal{H}(Y|\mathbf{x})} \forall k$ %
          \end{enumerate}
         \item Acquire labels for nearest-neighbor to centroids $X_{\mathcal{L}\target} ^{\timestep} \gets \{ \mathbf{NN}(\mathbf{\mu_{i}})\}_{i=1}^B$
         \item $X_{\mathcal{L}\target} = X_{\mathcal{L}\target} \cup X_{\mathcal{L}\target}^\timestep$
      \end{enumerate} 
    \State \textbf{Semi-supervised DA:} Update model $\modelparams^{\timestep+1}$. %
  \EndFor
  \State {\bf Return}: Final model parameters $\modelparams^{R+1}$.
  \caption{\method: Our proposed Active DA method, which uses Clustering Uncertainty-weighted Embeddings (\method) to select instances for labeling followed by a model update via semi-supervised domain adaptation.}
  \label{algo:training}
  \end{algorithmic}  
  \end{algorithm}

To jointly capture both diversity and uncertainty, we propose \emph{weighting samples based on their uncertainty} (given by Eq.~\ref{eq:ent}), and compute the weighted population variance~\cite{price1972extension}. 
The overall set-partitioning objective is:
\begin{equation}
  \begin{aligned}
    \argmin_{\setpartition, \mu} \sum_{\cluster=1}^{\numclusters} & \frac{1}{Z_k}\sum_{\mathbf{x} \in X_\cluster} \mathcal{H}(Y|\mathbf{x}) ||\featext(\mathbf{x}) - \mathbf{\mu_\cluster}||^2 
    \label{eq:clue}
  \end{aligned}
\end{equation}
where the normalization $Z_k=\sum_{x \in X_k} \mathcal{H}(Y|\mathbf{x})$. Our weighted set partitioning can also be viewed as standard set partitioning in an alternate feature space, where the density of instances is artificially increased proportional to their predictive entropy. Intuitively, this emphasizes representative sampling from uncertain regions of the feature space.

Since the objective in Eq.~\ref{eq:clue} is NP-hard, we approximate it using a Weighted K-Means algorithm~\cite{huang2005automated} (see Algorithm~\ref{algo:training} -- uncertainty-weighting is used in the update step). We set $K= \budget$ (budget), and use activations from the penultimate CNN layer as $\featext(\mathbf{x})$. After clustering, to select representative instances (\ie non-outliers), we acquire labels for the nearest neighbor to the weighted-mean of each set $\mathbf{\mu_k}$ in Eq.~\ref{eq:clue}.
Note that Eq.~\ref{eq:clue} equivalently maximizes the sum of squared deviations between instances in different sets~\cite{kriegel2017black} , ensuring that the constructed batch of instances has minimum redundancy.

\noindent \textbf{Trading-off uncertainty and diversity.} \method captures an implicit tradeoff between model uncertainty (via entropy-weighting) and feature-space coverage (via clustering). Consider the predictive probability distribution for instance $\mathbf{x}$: 

\vspace{-5pt}
\begin{equation}
p_{\modelparams}(Y | \mathbf{x}) = \sigma \left(\frac{\model(\mathbf{x})}{T}\right)
\end{equation}

\noindent where $\sigma$ denotes the softmax function and $T$ denotes its temperature. We observe that by modulating $T$, we can control the uncertainty-diversity tradeoff. For example, by increasing $T$, we obtain more diffuse softmax distributions for all points leading to similar uncertainty estimates across points; correspondingly, we expect diversity to play a bigger role. Similarly, at lower values of $T$ we expect uncertainty to have greater influence.

Our full label acquisition approach, Clustering Uncertainty-weighted Embeddings (\method), thus identifies instances that are both uncertain and diverse (see Fig.~\ref{fig:approach}). 

\noindent \textbf{Domain adaptation.} After acquiring labels via \method, we proceed to the next step of active adaptation: we update the model using the acquired target labels and optionally, the labeled source and unlabeled target data (see Fig.~\ref{fig:approach}, \emph{right}). In our main experiments (Sec.~\ref{subsec:results}), we experiment with 3 learning strategies: i) finetuning on target labels, ii) domain-adversarial learning via \texttt{DANN}~\cite{ganin2014unsupervised} with an additional target cross-entropy loss, and iii) semi-supervised adaptation via minimax entropy (\texttt{MME}~\cite{saito2019semi}). In Sec~\ref{subsec:learning} we also combine \method with additional DA methods from the literature.

Algorithm~\ref{algo:training} describes our full approach when using \method in combination with semi-supervised domain adaptation. Given a model trained on labeled source instances, we align its representations with unlabeled target instances via unsupervised domain adaptation. For $R$ rounds with per-round budget $\budget$, we iteratively i) acquire labels for $\budget$ target instances that are identified via our proposed sampling approach (\method), and ii) Update the model using a semi-supervised domain alignment strategy.

\vspace{-5pt}
\section{Experiments}
\vspace{-5pt}

We begin by describing our datasets and metrics, implementation details, and baselines (Sec~\ref{subsec:datasets}-~\ref{subsec:baselines}). Next, we benchmark the performance of \method across 6 domain shifts of varying difficulty against state-of-the art methods for Active DA and AL, across different learning settings (Sec~\ref{subsec:results}). We then ablate our method, analyze its sensitivity to various hyperparameters, and visualize its behavior (Sec~\ref{subsec:analysis}). Finally, we combine our method with various DA strategies, and study its effectiveness in learning from scratch (Sec~\ref{subsec:learning}). 
We follow the standard batch active learning setting~\cite{brinker2003incorporating}, in which we perform multiple rounds of batch active sampling, label acquisition, and model updates. %

\vspace{-5pt}
\subsection{Datasets and Metrics}
\label{subsec:datasets}

\noindent \textbf{DomainNet.} DomainNet~\cite{peng2019moment} is a large domain adaptation benchmark for image classification, containing 0.6 million images from 6 distinct domains spanning 345 categories.
We study four shifts of increasing difficulty as measured by source$\rightarrow$target transfer accuracy (TA): Real$\rightarrow$Clipart (easy, TA=40.6\%), Clipart$\rightarrow$Sketch (moderate, TA=34.7\%), Sketch$\rightarrow$Painting (hard, TA=30.3\%), and Clipart$\rightarrow$Quickdraw (very hard, TA=11.9\%). 

\noindent \textbf{DIGITS and Office.} We also report performance on the SVHN~\cite{netzer2011reading}$\rightarrow$MNIST~\cite{lecun1998gradient} and DSLR$\rightarrow$Amazon~\cite{saenko2010adapting} shifts.

\noindent \textbf{Metric.} We compute model accuracy on the target test split versus the number of labels used from the target train split at each round. We run each experiment 3 times and report mean accuracies. 
For clarity, we report performance at 3 randomly chosen intermediate budgets in the main paper and include full plots (mean accuracies and 1 standard deviation over all rounds) in the supplementary.
\vspace{-5pt}
\subsection{Implementation details}
\label{subsec:impl}
\vspace{-3pt}
\noindent \textbf{DomainNet.} We use a ResNet34~\cite{he2016deep} CNN, and perform 10 rounds of Active DA with a randomly selected per-round budget $= 500$ instances (total of 5000 labels). On DomainNet, we use the Clipart$\to$Sketch shift as a validation shift and use a small target validation set to select a softmax temperature of $T=0.1$ which we use for all other DomainNet shifts (details in supplementary). We include a sensitivity analysis over $T$ and $B$ in Sec.~\ref{subsec:analysis}. 

\noindent \textbf{DIGITS.} We match the experimental setting to Su~\etal~\cite{su2019active}: we use a modified LeNet architecture~\cite{hoffman2017cycada}, and perform 30 rounds of Active DA with B$=$10. 

\noindent \textbf{Office.} We use a ResNet34 CNN and perform 10 rounds of Active DA with B$=$30.  \noindent On DIGITS and Office, we use the default value of $T=1.0$. \noindent Across datasets, we use penultimate layer embeddings for \method and implement weighted K-means with $K=B$. All models are first trained on the labeled source domain. When adapting via semi-supervised domain adaptation, we additionally employ unsupervised feature alignment to the target domain at round 0. For additional details see supplementary.
\vspace{-5pt}
\subsection{Baselines}
\label{subsec:baselines}
\vspace{-3pt}
\noindent We compare \method against several state-of-the art methods for Active DA and Active Learning.

\noindent \textbf{1) \texttt{AADA}:} Active Adversarial Domain Adaptation~\cite{su2019active} (\texttt{AADA}) is a state-of-the-art Active DA method which performs alternate rounds of active sampling and adversarial domain adaptation via \texttt{DANN}~\cite{ganin2016domain}. It samples points with high predictive entropy and high probability of belonging to the target domain as predicted by the domain discriminator.

\noindent Further, we also benchmark the performance of 4 diverse AL strategies from prior work in the Active DA setting.

\noindent \textbf{2) \texttt{entropy}~\cite{wang2014new}:} Selects instances for which the model has highest predictive entropy. 

\noindent \textbf{3) \texttt{margin}~\cite{roth2006margin}:} Selects instances for which the score difference between the model's top-2 predictions is the smallest.

\noindent \textbf{4) \texttt{coreset}~\cite{sener2017active}:} Core-set formulates active sampling as a set-cover problem, and solves the K-Center~\cite{wolf2011facility} problem. We use the greedy version proposed in Sener et al.~\cite{sener2017active}.

\noindent \textbf{5) \texttt{BADGE}~\cite{ash2019deep}:} BADGE is a recently proposed state-of-the-art active learning strategy that constructs diverse batches by running KMeans++~\cite{arthur2006k} on ``gradient embeddings'' that incorporate model uncertainty and diversity. 

Methods (2) and (3) are uncertainty based, (4) is diversity-based, and (1) and (5) are hybrid approaches.

\begin{table*}
    \resizebox{\textwidth}{!}{
    \begin{tabular}{clcccccccccccccccc}
        \toprule
        \multirow{2}{*}{\centering DA method} & \multirow{2}{*}{\centering AL method} & {\centering AL} &\multicolumn{3}{c}{$\mathbf{R} \to \mathbf{C}$ (easy)} & \multicolumn{3}{c}{$\mathbf{C} \to \mathbf{S}$ (moderate)} & \multicolumn{3}{c}{$\mathbf{S} \to \mathbf{P}$ (hard)} & \multicolumn{3}{c}{$\mathbf{C} \to \mathbf{Q}$ (very hard)} & \multicolumn{3}{c}{AVG} \\
        & & {\centering Type} & \small{1k} & \small{2k}& \small{5k} & \small{1k} & \small{2k}& \small{5k} & \small{1k} & \small{2k}& \small{5k} & \small{1k} & \small{2k} & \small{5k} & \small{1k} & \small{2k} & \small{5k} \\ 
        \midrule            
        \multirow{6}{*}{\centering \makecell{\texttt{ft} \\ from \\ source}} & \texttt{uniform} & - & 51.5 & 55.3 & 60.6 & 42.1 & 44.4 & 47.0 & 41.1 & 43.8 & 47.2 & 23.3 & 28.1 & 35.3 & 39.5 & 42.9 & 47.5 \\
        & \texttt{entropy}~\cite{wang2014new} & U & 48.1 & 52.1 & 58.6 & 41.1 & 42.7 & 45.7 & 41.2 & 43.8 & 47.2 & 21.9 & 26.4 & 34.0 & 38.1 & 41.3 & 46.4 \\
        & \texttt{margin}~\cite{roth2006margin} & U & 51.0 & 54.8 & 60.7 & 42.3 & 44.3 & 47.0 & 41.4 & 44.0 & 47.1 & 23.6 & 28.4 & 35.8 & 39.6 & 42.9 & 47.7 \\
        & \texttt{coreset}~\cite{sener2017active} & D & 50.0 & 54.0 & 59.6 & 41.2 & 42.8 & 44.9 & 40.1 & 42.2 & 45.4 & 22.4 & 26.0 & 32.4 & 38.4 & 41.3 & 45.6 \\
        & \texttt{BADGE}~\cite{ash2019deep} & H & 52.4 & 56.1 & 61.7 & 42.8 & 45.2 & 48.1 & 41.7 & 44.9 & 47.9 & 23.1 & 28.2 & \textbf{35.5} & 39.8 & 43.6 & 48.3 \\
        \rowcolor{Gray}
        & \texttt{CLUE} (Ours) & H & \textbf{52.9} & \textbf{57.1} & \textbf{62.0} & \textbf{43.3} & \textbf{45.8} & \textbf{48.6} & \textbf{42.4} & \textbf{45.3} & \textbf{48.3} & \textbf{24.3} & \textbf{28.8} & \textbf{35.5} & \textbf{40.7} & \textbf{44.3} & \textbf{48.6} \\
        \midrule                    
        \multirow{6}{*}{\centering \makecell{\texttt{MME}~\cite{saito2019semi} \\ from \\ source}} & \texttt{uniform} & - & 55.2 & 59.3 & 63.5 & 45.7 & 47.8 & 49.7 & 42.9 & 45.3 & 47.8 & 24.5 & 30.3 & 38.1 & 42.1 & 45.7 & 49.8 \\
        & \texttt{entropy}~\cite{wang2014new} & U & 53.8 & 58.6 & 64.4 & 44.2 & 45.7 & 48.5 & 41.6 & 43.9 & 47.2 & 21.9 & 25.7 & 32.8 & 40.4 & 43.5 & 48.2 \\
        & \texttt{margin}~\cite{roth2006margin} & U & 55.6 & \textbf{60.7} & \textbf{65.7} & 46.0 & 48.1 & 50.8 & 42.2 & 44.8 & 48.2 & 23.1 & 28.3 & 36.6 & 41.7 & 45.5 & 50.3 \\
        & \texttt{coreset}~\cite{sener2017active} & D & 54.3 & 59.1 & 64.6 & 45.1 & 46.7 & 48.9 &  42.4 & 44.2 & 47.1 & 23.9 & 27.8 & 34.3 & 41.4 & 44.5 & 48.7 \\
        & \texttt{BADGE}~\cite{ash2019deep} & H & 56.2 & 60.6 & \textbf{65.7} & 45.8 & 48.2 & 50.7 & 43.1 & 45.7 & 48.7 & 24.3 & 29.6 & 38.3 & 42.4 & 46.0 & 50.9 \\
        \rowcolor{Gray}
        & \texttt{CLUE} (Ours) & H & \textbf{56.3} & \textbf{60.7} & 65.3 & \textbf{46.8} & \textbf{49.0} & \textbf{51.4} & \textbf{43.7} & \textbf{46.5} & \textbf{49.4} & \textbf{25.6} & \textbf{31.1} & \textbf{38.9} & \textbf{43.1}  & \textbf{46.8} & \textbf{51.3} \\
        \midrule        
        \texttt{DANN}~\cite{ganin2014unsupervised} & \texttt{AADA}~\cite{su2019active} & H & 53.2 & 57.4 & 62.8 & 44.8 & 46.5 & 49.2 & 41.3 & 43.5 & 46.1 & 21.9 & 25.8 & 32.4 & 40.3 & 43.3 & 47.6 \\
        \rowcolor{Gray}
        from source &  \method (Ours) & H & \textbf{54.6} & \textbf{58.9} & \textbf{63.8} & \textbf{45.3} & \textbf{47.9} & \textbf{50.8} & \textbf{43.2} & \textbf{45.5} & \textbf{48.3} & \textbf{24.4} & \textbf{29.2} & \textbf{35.4} & \textbf{41.9} & \textbf{45.4} & \textbf{49.6} \\
        \bottomrule
        \end{tabular}}
        \vspace{-2pt}
        \caption{Accuracies on target test set for 4 DomainNet shifts of increasing difficulty spanning 5 domains: Real (R), Clipart (C), Sketch (S), Painting (P) and Quickdraw (Q). We perform 10 rounds of Active DA with $B=500$ and report results at 3 intermediate rounds (full plots in supplementary), as well as the 4-shift average (AVG). We compare \method against state-of-the art methods for AL (\texttt{entropy}~\cite{wang2014new}, \texttt{margin}~\cite{roth2006margin}, \texttt{coreset}~\cite{sener2017active}, \texttt{BADGE}~\cite{ash2019deep}) and Active DA (\texttt{AADA}), spanning different AL paradigms: uncertainty sampling (U), diversity sampling (D), and hybrid (H) combinations of the two. We use multiple learning strategies: finetuning (\texttt{ft}), \texttt{MME}~\cite{saito2019semi} (state-of-the-art semi-supervised DA method), and semi-supervised DA via \texttt{DANN}~\cite{ganin2014unsupervised}. Best performance is in bold, gray rows are our method.
            }\vspace{-5pt}
        \label{fig:domainnet}
\end{table*}

\begin{table}
    \setlength{\tabcolsep}{4pt}
    \resizebox{\textwidth}{!}{
    \begin{tabular}{clcccccc}
        \toprule
        \multirow{2}{*}{\centering DA method} & \multirow{2}{*}{\centering AL method} &\multicolumn{3}{c}{$\mathbf{SVHN} \to \mathbf{MNIST}$} & \multicolumn{3}{c}{$\mathbf{DSLR} \to \mathbf{Amazon}$} \\
        & & \small{30} & \small{60}& \small{150} & \small{30} & \small{60}& \small{150} \\ 
        \midrule            
        \multirow{6}{*}{\centering \makecell{\texttt{ft} \\ from \\ source}} & \texttt{uniform} & 77.7 & 88.2 & 95.2 & 54.3 & 58.0 & 67.5 \\
        & \texttt{entropy}~\cite{wang2014new} & 65.8 & 75.6 & 92.9 & 51.2 & 52.4 & 59.1  \\
        & \texttt{margin}~\cite{roth2006margin} & 82.0 & 89.3 & \textbf{95.5}        & 52.4 & 54.4 & 65.5  \\
        & \texttt{coreset}~\cite{sener2017active} & 71.6 & 76.5 & 87.9 & 53.9 & 55.8 & 67.2  \\
        & \texttt{BADGE}~\cite{ash2019deep} & 78.7 & 88.2 & 95.2 & 55.8 & 59.2 & \textbf{71.0} \\
        \rowcolor{Gray}
        & \texttt{CLUE} (Ours)& \textbf{83.9} & \textbf{89.4} & 94.5 & \textbf{56.4} & \textbf{60.5} & 70.5 \\
        \midrule                    
        \multirow{6}{*}{\centering \makecell{\texttt{MME}~\cite{saito2019semi} \\ from \\ source}} & \texttt{uniform} & 85.5 & 91.2 & 95.0 & 58.3 & 61.7 & 70.0  \\
        & \texttt{entropy}~\cite{wang2014new} & 81.3 & 85.7 & 93.9 & 54.9 & 56.5 & 66.2  \\
        & \texttt{margin}~\cite{roth2006margin} & 88.4 & 91.5 & 96.6 & 54.7 & 58.5 & 70.6  \\
        & \texttt{coreset}~\cite{sener2017active} & 85.8 & 89.1 & 94.6 & 57.7 & 61.0 & 70.5  \\
        & \texttt{BADGE}~\cite{ash2019deep} & 89.9 & 93.1 & \textbf{96.4} & 58.2 & 61.6 & 71.3  \\
        \rowcolor{Gray}
        & \texttt{CLUE} (Ours) & \textbf{91.1} & \textbf{93.9} & 96.2 & \textbf{60.2} & \textbf{65.6} & \textbf{72.7}  \\
        \midrule                    
        \texttt{DANN}~\cite{ganin2014unsupervised} & \texttt{AADA}~\cite{su2019active} & 88.8 & 90.7 & \textbf{95.4} & 54.2 & 56.6 & 65.4  \\        
        \rowcolor{Gray} 
        from source & \method (Ours) & \textbf{90.9} & \textbf{93.1} & 95.3 & \textbf{59.1} & \textbf{64.5} & \textbf{72.1}  \\
        \bottomrule
        \end{tabular}}
        \vspace{-5pt}
        \caption{Active DA accuracies on target test set at 3 intermediate budgets (30, 60, 150) for: \textbf{Middle:} 30 rounds with $B=10$ from SVHN$\to$MNIST (DIGITS). \textbf{Right:} 10 rounds with $B=30$ from  DSLR$\to$Amazon (Office). Best performance is in bold, gray rows are our method. 
        For full plots see supplementary.
        }\vspace{-5pt}
        \label{fig:digits_office}
\end{table}

\vspace*{-20pt}
\subsection{Results}
\label{subsec:results}
\vspace{-3pt}
\noindent We evaluate all methods across three ways of learning in the presence of a domain shift with the acquired labels:

\noindent 1) \textbf{\texttt{FT} from source}: Finetuning a model trained on the source domain with acquired target labels.

\noindent 2) \textbf{\texttt{MME}~\cite{saito2019semi} from source}: Minimax entropy~\cite{saito2019semi} (\texttt{MME}) is a state-of-the-art semi-supervised DA method that starts from a source model and minimizes an adversarial entropy loss for unsupervised domain alignment in addition to finetuning on labeled source and target data.

Tables~\ref{fig:domainnet} and ~\ref{fig:digits_office} demonstrate our results on DomainNet, DIGITS, and Office. We make the following observations:

\vspace{-2pt}
\noindent $\triangleright$ \textbf{Uncertainty and diversity sampling are less effective for Active DA, frequently underperforming even random sampling.} Approaches solely based on uncertainty (\eg \texttt{margin}~\cite{roth2006margin}) work well on relatively easier shifts (R$\to$C with \texttt{MME}, SVHN$\to$MNIST), but overall we find uncertainty-based (\texttt{margin}, \texttt{entropy}), and diversity-based (\texttt{coreset}) approaches generalize poorly to challenging shifts (\eg S$\to$P, C$\to$Q), frequently underperforming even random sampling! On the other hand, hybrid approaches (\method and \texttt{BADGE}) that combine uncertainty and diversity are versatile across shift difficulties.

\vspace{-3pt}
\noindent $\triangleright$ \textbf{\method outperforms prior AL methods in the Active DA setting.} Across learning strategies, shifts, benchmarks, and most rounds, \method consistently performs best. 
Averaged over 4 DomainNet shifts, \method outperforms 
\texttt{margin}-based uncertainty sampling and \texttt{coreset}-based diversity sampling at $B=2k$ by 1.4\% and 3\% when finetuning, and 1.3\% and 2.3\% when adapting via \texttt{MME} (Tab.~\ref{fig:domainnet}). Similarly, \method outperforms the next best-performing method (\texttt{BADGE}) by 0.7\% at $B=2k$ when finetuning and 0.8\% with \texttt{MME} (4 shift average). While \texttt{BADGE}~\cite{ash2019deep} is also a hybrid AL method that combines uncertainty and diversity sampling, it does so by clustering in a high-dimensional ``gradient-embedding'' space ($\sim$176k dimensions on C$\to$S with a ResNet34, versus 512-dimensional embeddings used by \method, details in Tab.~\ref{fig:time}), in which distance-based diversity measures may be less meaningful due to the curse of dimensionality. We note here that DomainNet is a complex benchmark with 345 categories and significant label noise, which often leads to relatively small absolute margins of improvement; however, our results demonstrate \method's versatility in generalizing across diverse shifts without shift-specific tuning. 

On DIGITS and Office, \method's gains are even more significant (Tab.~\ref{fig:digits_office}). For instance at $B=30$ on SVHN$\to$MNIST, \method improves upon \texttt{margin}, \texttt{coreset}, and \texttt{BADGE} when finetuning by 1.9\%, 12.3\%, and 5.2\%, and 4\%, 2.5\% and 0.6\% on DSLR$\to$Amazon. 

\begin{figure*}[t]
    \centering
    \includegraphics[width=\textwidth]{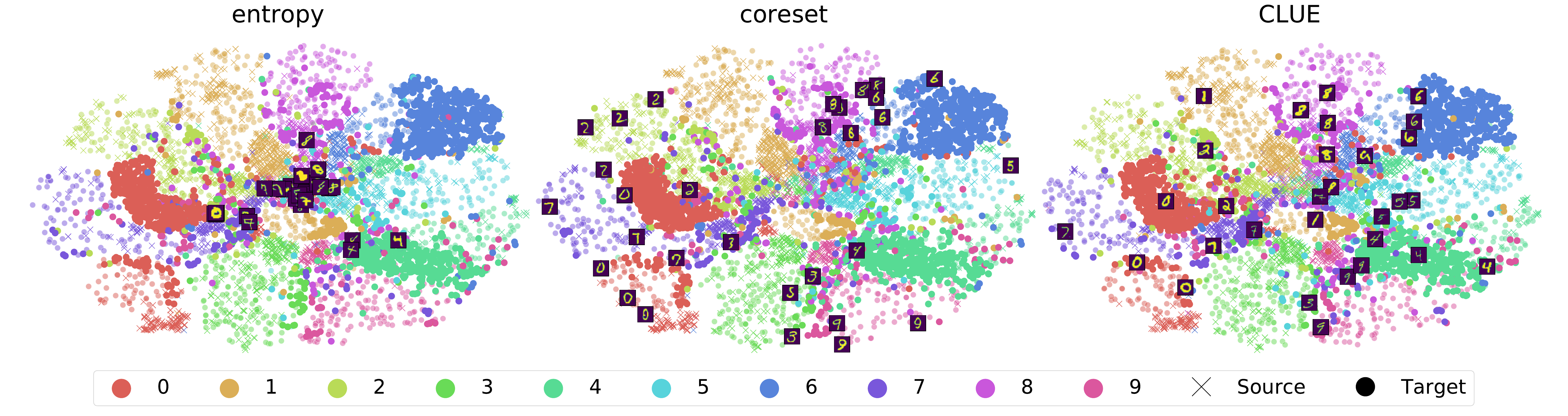}
    \vspace*{-15pt}
    \caption{SVHN$\rightarrow$MNIST: We visualize the logits of a subset of incorrect (large, opaque circles) and correct (partly transparent circles) model predictions on the target domain after round 0, along with examples sampled by different methods. \texttt{entropy}~\cite{wang2014new} (\emph{left}) acquires redundant samples, whereas \texttt{coreset}~\cite{sener2017active} (\emph{middle}) does not account for areas of the feature space that are already well-aligned across domains. \method (\emph{right}) constructs batches of dissimilar samples from dense regions with high uncertainty.}
    \vspace*{-5pt}
    \label{fig:tsne}
\end{figure*}

\noindent $\triangleright$ \textbf{Additional unsupervised adaptation helps in the Active DA setting.} Across AL methods, we observe adaptation with \texttt{MME} to consistently outperform finetuning (\eg by 2.4-2.7\% accuracy on DomainNet). 

\noindent $\triangleright$ \textbf{\method significantly outperforms the state-of-the art Active DA method \texttt{AADA}.} \texttt{AADA}~\cite{su2019active} acquires labels by using a domain classifier learned via \texttt{DANN}~\cite{ganin2014unsupervised}. Thus, it is undefined in the \texttt{FT} and \texttt{MME} settings. For an apples-to-apples comparison, we report performance of \method+\texttt{DANN} in the last 2 rows of Tabs.~\ref{fig:domainnet} and ~\ref{fig:digits_office}. As seen, \method+\texttt{DANN} consistently outperforms \texttt{AADA}, the state-of-the art Active DA method, \eg by 0.4\%-2\% on DomainNet. Further, we find the performance gap between our method and \texttt{AADA}~\cite{su2019active} increases with increasing shift difficulty, as predictive uncertainty becomes increasingly unreliable (3.4\% gain at $B=2k$ on the very hard C$\rightarrow$Q shift). We observe similar improvements over \texttt{AADA} on the DIGITS and Office (Tab.~\ref{fig:digits_office}) benchmarks, \eg 2.4\% and 7.9\% at $B=60$. Further, our best performing \method + \texttt{MME} strategy improves the gains still further to as much as 3.5\% at $B=2k$ on DomainNet and 9\% at $B=60$ on Office!

As discussed in Sec.~\ref{sec:approach}, the optimal label acquisition criterion may vary across shifts and stages of training as the model's uncertainty estimates and feature space evolve, and it is challenging for a single approach to work well consistently. Despite this, \method effectively trades-off uncertainty and diversity to generalize reliably across shifts. 
    
\subsection{Analyzing and Ablating \method}
\label{subsec:analysis}

\begin{figure*}[h]
  \centering
  \begin{subfigure}[b]{0.23\textwidth}
    \centering
    \includegraphics[width=1.0\linewidth]{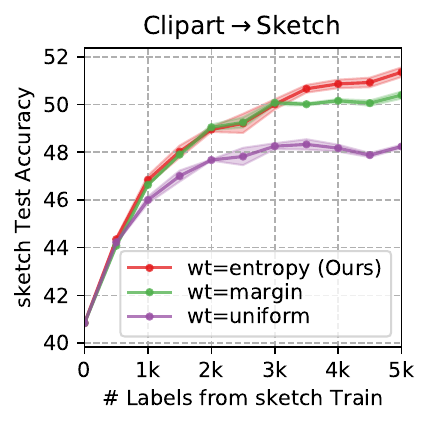}
    \caption[]%
    {{\small Varying uncertainty measure in \method.}}
    \label{fig:cs_clue_ablate}
    \end{subfigure}\hspace{0.5em}
    \begin{subfigure}[b]{0.23\textwidth}
      \centering
      \includegraphics[width=1.0\linewidth]{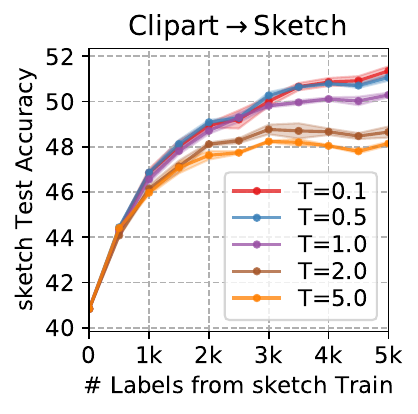}
      \caption[]%
      {{\small Measuring \method's sensitivity to temperature T.}}
      \label{fig:cs_temp}
      \end{subfigure}\hspace{0.5em}
  \begin{subfigure}[b]{0.23\textwidth}
      \includegraphics[width=1.0\linewidth]{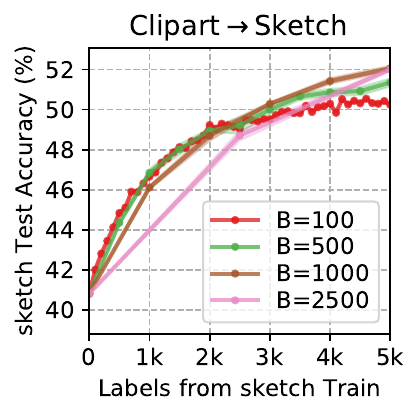}
      \caption[]%
      {{\small Measuring \method's sensitivity to budget B.}}
      \label{fig:cs_budget}
  \end{subfigure}\hspace{0.5em}
  \begin{subfigure}[b]{0.23\textwidth}
  \centering
  \includegraphics[width=1.0\textwidth]{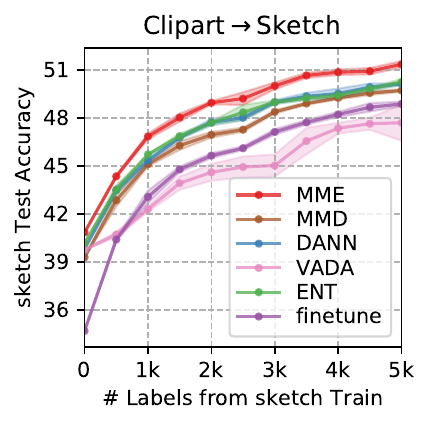}
  \caption[]%
  {{\small Varying DA method while sampling via \method.}}  
  \label{fig:cs_train_ablate}
  \end{subfigure}
  \vspace{-2pt}
  \caption{\textbf{(a), (b), (c)}: Ablating and analyzing \method on C$\rightarrow$S. \textbf{(d)}: Combining \method with different DA strategies on C$\to$S. Best viewed in color. We perform 10 rounds of Active DA with B=500, and report accuracy mean and 1 standard deviation (via shading) over 3 runs.}
  \vspace{-6pt}
  \label{fig:hyperparams}
\end{figure*}

\noindent \textbf{Visualizing \method via t-SNE.} We provide an illustrative comparison of sampling strategies using t-SNE~\cite{maaten2008visualizing}. Fig.~\ref{fig:tsne} shows an initial feature landscape together with points selected by \texttt{entropy}-based uncertainty sampling, diversity-based \texttt{coreset} sampling, and \method at Round 0 on the SVHN$\rightarrow$MNIST shift. We find that \texttt{entropy}~\cite{wang2014new} (\textit{left}) samples uncertain but redundant points, \texttt{coreset}~\cite{sener2017active} samples diverse but not necessarily uncertain points, while our method, \method, samples both diverse and uncertain points. In the supplementary, we include visualizations over several rounds and find that \method consistently selects diverse target instances from dense, uncertain regions of the feature space.

\noindent \textbf{Varying uncertainty measure in \method}. In Fig.~\ref{fig:cs_clue_ablate}, we consider alternative uncertainty measures for \method on the C$\rightarrow$S shift. We show that our proposed use of sample entropy significantly outperforms a uniform sample weight and narrowly outperforms an alternative uncertainty measure - sample margin score (difference between scores for top-2 most likely classes). This illustrates the importance of using uncertainty-weighting to bias \method towards informative samples. 
We also experimented (not shown) with last-layer embeddings (instead of penultimate) for \method, and observed near-identical performance across multiple shifts, suggesting that \method is not sensitive to this choice.

\noindent \textbf{Sensitivity to parameters.} In Fig.~\ref{fig:hyperparams} we measure \method's sensitivity to two parameters: the softmax temperature hyperparameter T and experimental parameter budget B.

\noindent \textbf{i) Sensitivity to softmax temperature T.} %
Recall from Sec.~\ref{sec:approach} that by tuning the softmax temperature in \method, we can vary the trade-off between uncertainty and diversity. In Fig.~\ref{fig:cs_temp} we run a sweep over temperature values used for \method on C$\rightarrow$S . As seen, lower values of temperature (which emphasizes the role of uncertainty) improve performance, particularly at later rounds when uncertainty estimates are more reliable. We note that $T$ is an optional hyperparameter that may be tuned if a small target validation set is available, but \method obtains strong state-of-the art results across across DIGITS, Office, and DomainNet even with the default value of T=1.0.
On DomainNet, we further improve performance by selecting $T=0.1$ via grid search on a single C$\to$S shift and find that it generalizes to other DomainNet shifts.

\noindent  \textbf{ii) Sensitivity to budget B.} We now vary the per-round budget (and consequently the total number of active adaptation rounds) and report performance on the Clipart$\rightarrow$Sketch shift. As seen in Fig.~\ref{fig:cs_budget}, \method performs well across budget values of 100, 500, 1k and 2.5k. We also observe consistent performance with a different budget ($B=30$) on the SVHN$\to$MNIST shift (details in supplementary).

\noindent \textbf{Time complexity}. Table~\ref{fig:time} shows the average case complexity and AL querying time-per-round on SVHN$\to$MNIST and C$\to$S. \texttt{CLUE} and \texttt{BADGE}, which achieve the best accuracy, are slower to run due to a (CPU) clustering step. \texttt{CLUE} can be optimized further via GPU acceleration, using last-(instead of penultimate) layer embeddings, or pre-filtering data before clustering.

\begin{table}[h]
  \vspace{-7pt}
    \begin{tabular}{llcc} \toprule
        & \begin{tabular}[c]{@{}c@{}}AL\\ Strategy\end{tabular} & \begin{tabular}[c]{@{}c@{}}Query\\ Complexity\end{tabular}      & \begin{tabular}[c]{@{}c@{}}Query Time\\ { \small(DIGITS, C$\rightarrow$S)}\end{tabular} \\ \hline
        \multirow{3}{*}{\rotatebox{90}{\centering \begin{tabular}[c]{@{}c@{}}fwd + \\cluster\end{tabular} }} 
        &\texttt{CLUE} (Ours) &$\mathcal{O}(tNBD)$ & (60s, 16.2m)\\ 
        &\texttt{BADGE}~\cite{ash2019deep} & $\mathcal{O}(NBDC)$ & (103s, 16.3m) \\
        &\texttt{coreset}~\cite{sener2017active} & $\mathcal{O}(CNB)$ & (52s, 2.8m)\\
                \midrule
          \multirow{3}{*}{\rotatebox{90}{\centering \begin{tabular}[c]{@{}c@{}}fwd + \\rank\end{tabular} }}
        &\texttt{AADA}~\cite{su2019active}  & $\mathcal{O}(NC)$& (3.7s, 139s) \\
          &\texttt{entropy}~\cite{wang2014new}  &$\mathcal{O}(NC)$ & (3.5s, 45s)\\
          &\texttt{margin}~\cite{roth2006margin}  &$\mathcal{O}(NC)$& (3.2s, 45s)\\
                \bottomrule      
                \end{tabular}
                \vspace*{-5pt}
                \caption{Query complexity and time-per-round for \method and prior AL strategies. $C$ and $N$ denotes number of classes and instances, $D$ denotes embedding dimensionality, $B$ denotes budget, $t=$ number of clustering iterations, and fwd stands for forward pass.}
                \label{fig:time}
\end{table}

\vspace*{-5pt}
\subsection{\method across learning strategies}
\label{subsec:learning}
\vspace{-2pt}

\noindent \textbf{\method with different DA strategies.} We now study \method's compatibility with a few additional domain adaptation strategies from the literature. In addition to the finetuning, \texttt{DANN}~\cite{ganin2014unsupervised} and \texttt{MME}~\cite{saito2019semi} already studied in Sec.~\ref{subsec:results}, we fix our sampling strategy to \method and vary the learning strategy to: i) \texttt{MMD}~\cite{long2013transfer}, a discrepancy-statistic based DA method,  and ii) \texttt{VADA}~\cite{shu2018dirt}, a domain-classifier based method that uses virtual adversarial training. iii) \texttt{ENT}~\cite{grandvalet2005semi}: A variant of the \text{MME} method using standard rather than adversarial entropy minimization. Initial performance varies across methods since we employ unsupervised DA at Round 0.

In Fig.~\ref{fig:cs_train_ablate}, we observe that domain alignment with \texttt{MME} significantly outperforms all alternative methods. With all DA methods except \texttt{VADA}, we observe improvements over finetuning; however, \texttt{MME} clearly performs best. The improvements over \texttt{DANN} and \texttt{VADA} are consistent with Saito~\etal.~\cite{saito2019semi}, who find that domain-classifier based methods are less effective when some target labels are available.

\noindent \textbf{How well does \method learn from scratch?} While \method is designed for active learning under a domain shift, for completeness we also evaluate its performance against prior work when learning from ``scratch'' as is conventional in AL. We find that it outperforms prior work when finetuning using ImageNet~\cite{russakovsky2015imagenet} initialization on C$\rightarrow$S, and performs on par with competing methods when finetuning from scratch on SVHN~\cite{netzer2011reading} (details in supplementary).

\vspace{-5pt}
\section{Conclusion}
\label{sec:conclusion}
\vspace{-3pt}

We address active domain adaptation, where the goal is to select target instances for labeling so as to generalize a trained source model to a new target domain. We show how existing active learning strategies based solely on uncertainty or diversity sampling are not effective for Active DA.
We present \method, a novel label acquisition strategy for active sampling under a domain shift, that performs uncertainty-weighted clustering to select diverse, informative target instances for labeling from dense regions of the feature space. We demonstrate \method's effectiveness over competing active learning and Active DA methods across learning settings and domain shifts, and comprehensively analyze its behavior.

\noindent\textbf{Acknowledgements.} This work was supported by the DARPA LwLL program. We would like to thank Devi Parikh for guidance, and Prithvijit Chattopadhyay, Cornelia Köhler, and Shruti Venkatram for feedback on the draft.
{\small
\bibliographystyle{ieee_fullname}
\bibliography{main}
}

\section{Appendix}
\localtableofcontents

\newcommand\DoToC{%
  \startcontents
  \printcontents{}{2}{\textbf{Contents}\vskip3pt\hrule\vskip5pt}
  \vskip3pt\hrule\vskip5pt
}
\subsection{Performance of \method on Standard AL}
\label{sec:al_svhn}

While \method is designed as an Active DA strategy, we nevertheless study its suitability for traditional active learning in which models are typically trained from ``scratch''. We benchmark its performance against competing methods in two settings: Finetuning from an ImageNet~\cite{russakovsky2015imagenet} initialization on the Clipart$\rightarrow$Sketch shift from DomainNet, following the same experimental protocol we use for Active DA (but set softmax temperature T$=1.0$ as uncertainty is less reliable when learning from scratch), and the standard SVHN benchmark used for active learning~\cite{ash2019deep}. For AL on SVHN, we match the setting in Ash~\etal~\cite{ash2019deep}, initializing a ResNet18~\cite{he2016deep} CNN with random weights and perform 100 rounds of active learning with per-round budget of 100. As summarized in Figure~\ref{fig:al_c2s}, on DomainNet \method significantly outpeforms prior work. On SVHN (Fig~\ref{fig:results_al_svhn}), \method is on-par with state-of-the-art AL methods, and statistically significantly better than uniform sampling over most rounds.

\subsection{Analyzing \method: When do gains saturate?}

\begin{figure}[h]
    \centering
    \includegraphics[width=0.48\linewidth]{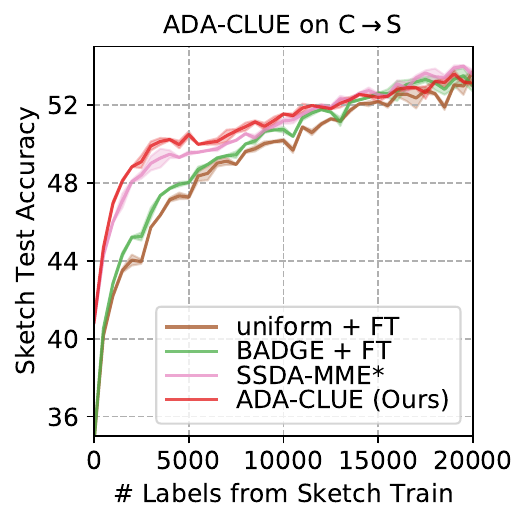}
    \caption{C$\rightarrow$S: When do gains with \method saturate?}
    \label{fig:cs_conv}
\end{figure}

\begin{figure*}[t]
    \centering
    \begin{subfigure}[b]{0.3\linewidth}
        \centering 
        \includegraphics[width=\textwidth]{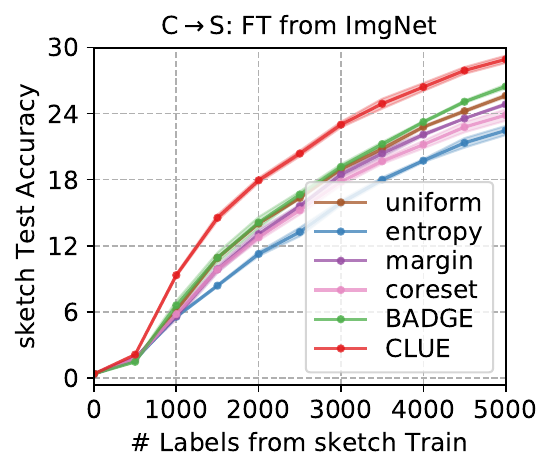}
        \caption[]%
        {{\small C$\rightarrow$S: 10 rounds,B $=500$}}   
        \label{fig:al_c2s}
    \end{subfigure}
    \begin{subfigure}[b]{0.65\linewidth}
        \centering
        \includegraphics[width=\textwidth]{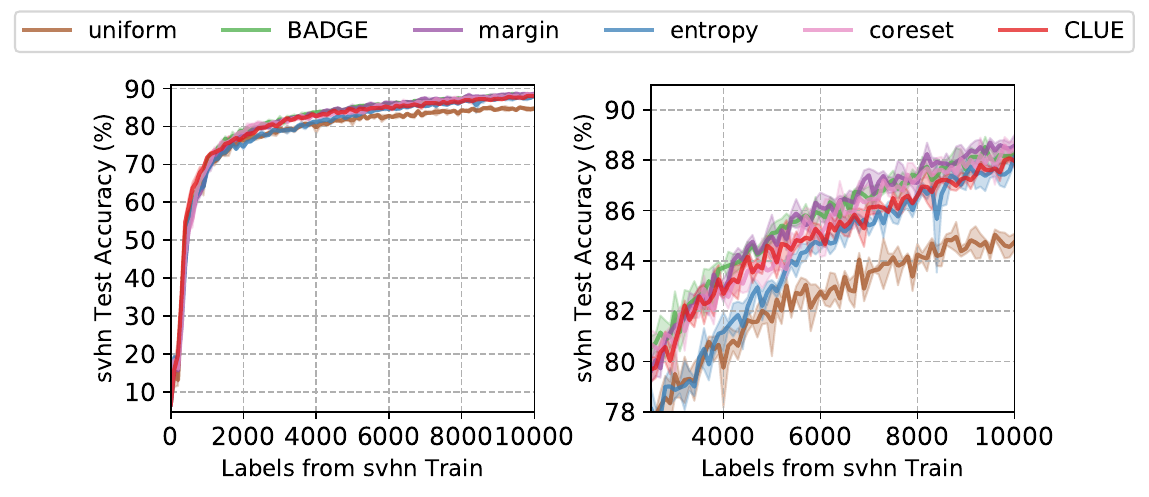}
        \caption[]%
        {{\small SVHN: 100 rounds, B$=100$}}
        \label{fig:results_al_svhn}
    \end{subfigure}
    \caption[]
    {Active learning performance of \method on DomainNet C$\rightarrow$S (finetuning from ImageNet initialization) and SVHN (finetuning from scratch). \method significantly outperforms state-of-the art active learning methods in the first case and performs on-par in the second.}
    \label{fig:active_learning}
\end{figure*}

\noindent Due to the computational expense of running active domain adaptation on multiple shifts with a large CNN (ResNet34~\cite{he2016deep}) on a large dataset (DomainNet~\cite{peng2019moment}), in the main paper we restrict ourselves to 10 rounds with a per-round budget of 500. As a check on when performance gains saturate, we benchmark performance of \method + \texttt{MME}~\cite{saito2019semi} on Clipart$\rightarrow$Sketch for 40 rounds with per-round budget of 500 ($= 20k$ labels in total), against a few representative baselines: \texttt{BADGE}~\cite{ash2019deep}+\texttt{FT} (state-of-the-art AL method with finetuning), \texttt{uniform}+\texttt{MME}~\cite{saito2019semi} (uniform sampling with state-of-the-art semi-supervised DA method), and \texttt{uniform} + \texttt{FT} (uniform sampling with finetuning). Results are presented in Fig.~\ref{fig:cs_conv}. As seen, \method's strongest performance improvements are seen in the initial stages of training, where it significantly outperforms competing methods. Performance then begins to saturate roughly around the 15k labels mark, and performance differences across methods narrow.

\subsection{Comparing \method and \texttt{BADGE}}
\label{sec:dset_details}

\noindent Both \method and \texttt{BADGE} are hybrid label acqusition strategies that combine uncertainty and diversity sampling. We now elaborate upon the differences between the two:

\noindent \textbf{Conceptual comparison}. \method and \texttt{BADGE} \emph{conceptually} differ in 3 key ways:  \textbf{i) Feature space}. \texttt{BADGE}: Operates in ``gradient embedding'' space computed as an outer product of penultimate-layer instance embeddings and model output scores. \method: Operates on penultimate-layer embeddings scaled by model uncertainty. \textbf{ii) Uncertainty measure}. \texttt{BADGE}: Uses gradient wrt model's top-1 prediction. \method: Uses predictive entropy. \textbf{iii) Diversity measure}. \texttt{BADGE}: Runs KMeans++ in gradient embedding space. \method: Runs uncertainty-weighted K-Means on instance embeddings and selects nearest neighbors to centroids.

These design choices influence both \texttt{BADGE}'s effectiveness and efficiency. For $D-$dim. penultimate-layer embeddings and $C-$dim. (for $C$ classes) output scores, \emph{each} gradient embedding has $CD$-dims -- on DomainNet, this is $176k$-dims with a ResNet34 and $\sim$1.4 million dims with AlexNet. In addition to being expensive to compute, KMeans++ in such high dimensional spaces is less effective as distance measures becomes less reliable. In comparison, \method operates in a significantly lower-dimensional feature space (512/4096 for ResNet34/AlexNet). We believe these differences lead to \method being more computationally efficient (Tab. 3 in main paper) and effective on average than \texttt{BADGE} (Tab. 1-2 in main paper), especially on hard shifts (\eg S$\to$P, C$\to$Q in Tab. 1 of main paper).

\begin{table}[h]
    \setlength{\tabcolsep}{4pt}
    \resizebox{\textwidth}{!}{
    \begin{tabular}{lcccccc}
        \toprule
        \multirow{2}{*}{\centering AL method} &\multicolumn{3}{c}{$\mathbf{SVHN} \to \mathbf{MNIST}$} & \multicolumn{3}{c}{$\mathbf{DSLR} \to \mathbf{Amazon}$} \\
        & \small{30} & \small{60}& \small{150} & \small{30} & \small{60}& \small{150}\\ 
        \midrule
        \texttt{BADGE} & \textbf{89.9{\scriptsize$\pm 0.9$}} & 93.1{\scriptsize$\pm 0.2$} &  \textbf{96.4{\scriptsize$\pm 0.1$}} & \textbf{58.2{\scriptsize$\pm 1.2$}} & 61.6{\scriptsize$\pm 0.2$} & 71.3{\scriptsize$\pm 1.4$} \\
        \texttt{CLUE} & \textbf{91.1{\scriptsize$\pm 0.5$}} & \textbf{93.9{\scriptsize$\pm 0.5$}} &  \textbf{96.2{\scriptsize$\pm 0.2$}} &  \textbf{60.2{\scriptsize$\pm 1.2$}} & \textbf{65.6{\scriptsize$\pm 0.5$}} & \textbf{72.7{\scriptsize$\pm 0.9$}} \\     
        \bottomrule
        \end{tabular}}
        \vspace{-5pt}
        \caption{{\small Active DA acc. and 1 std dev over 3 runs with \texttt{MME}.
        }}\vspace{-5pt}
        \label{tab:s2md2a}
  \end{table}

\noindent \textbf{Empirical comparison}. On DomainNet, \method achieves small but \emph{consistent} gains over \texttt{BADGE} (Tab. 1 in main paper) -- with \texttt{MME}, across 4 diverse shifts $\times$ 10 rounds = $40$ settings, and accounting for error bars, \method does as well or better than \texttt{BADGE} on 38/40 (better on 24/40). On other benchmarks, we sometimes observe large gains on other benchmarks (+5.2\% on DIGITS at B=30, Tab. 2). However, at later rounds on SVHN$\to$MNIST and DSLR$\to$Amazon, \texttt{BADGE} and \method perform similarly. With \texttt{MME}, and after including error bars, \method matches or outperforms {BADGE} on 24/30 (DIGITS) and 10/10 (Office) settings (Tab.~\ref{tab:s2md2a} shows 3 budgets), with \texttt{BADGE} doing slightly better from rounds 20-24 on DIGITS. We conjecture this is because gradient embedding dimensionality on DIGITS for 10-way classification is low, which leads to \texttt{BADGE} being as effective.

\subsection{Dataset details}
\label{sec:dset_details}

\begin{figure}[h]
    \centering
    \begin{subfigure}[b]{0.48\linewidth}  
        \centering 
        \includegraphics[width=\textwidth]{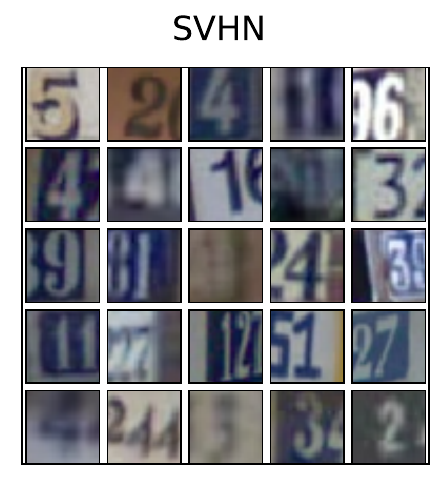}
        \caption[]%
        {{}}   
        \label{fig:svhn}
    \end{subfigure}
    \begin{subfigure}[b]{0.48\linewidth}
        \centering
        \includegraphics[width=\textwidth]{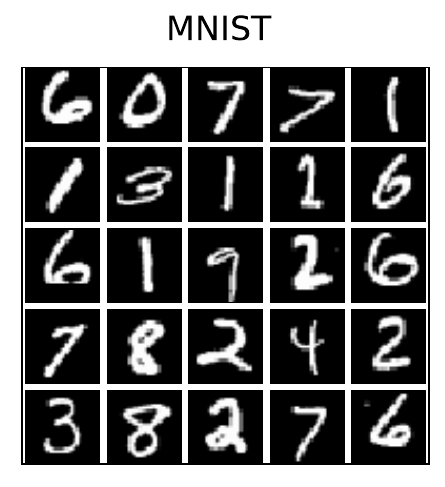}
        \caption[]%
        {{}}
        \label{fig:mnist}
    \end{subfigure}
    \caption[]
    {DIGITS qualitative examples}
    \label{fig:digits_qual}
\end{figure}

\begin{figure*}[b]
    \centering
    \begin{subfigure}[b]{0.19\textwidth}  
        \centering 
        \includegraphics[width=\textwidth]{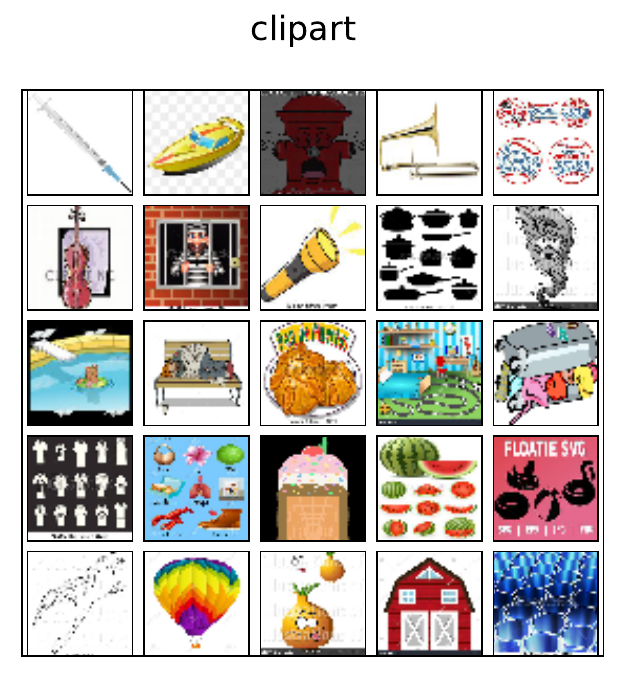}
        \caption[]%
        {{}}   
        \label{fig:clipart}
    \end{subfigure}
    \begin{subfigure}[b]{0.19\textwidth}
        \centering
        \includegraphics[width=\textwidth]{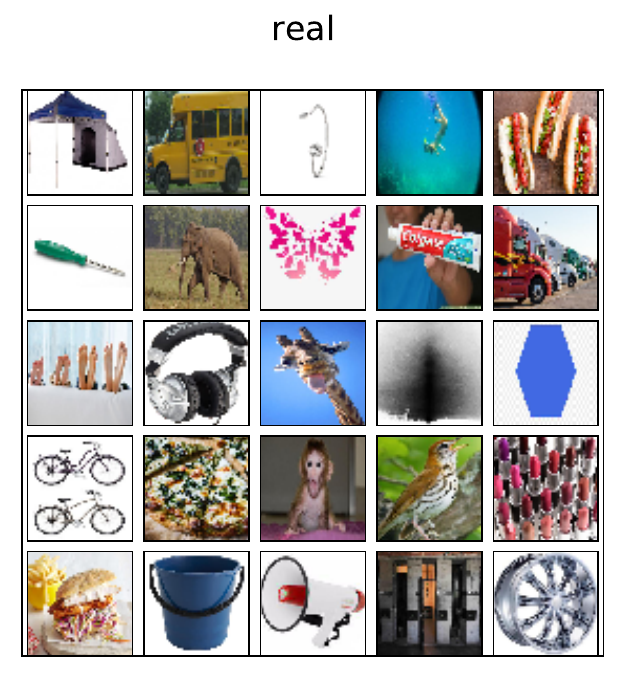}
        \caption[]%
        {{}}
        \label{fig:real}
    \end{subfigure}
    \begin{subfigure}[b]{0.19\textwidth}
        \centering
        \includegraphics[width=\textwidth]{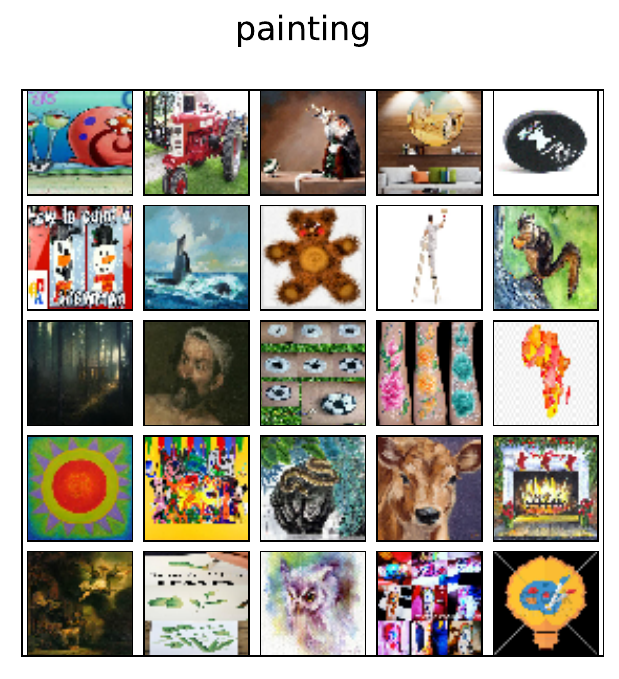}
        \caption[]%
        {{}}
        \label{fig:painting}
    \end{subfigure}
    \begin{subfigure}[b]{0.19\textwidth}
        \centering
        \includegraphics[width=\textwidth]{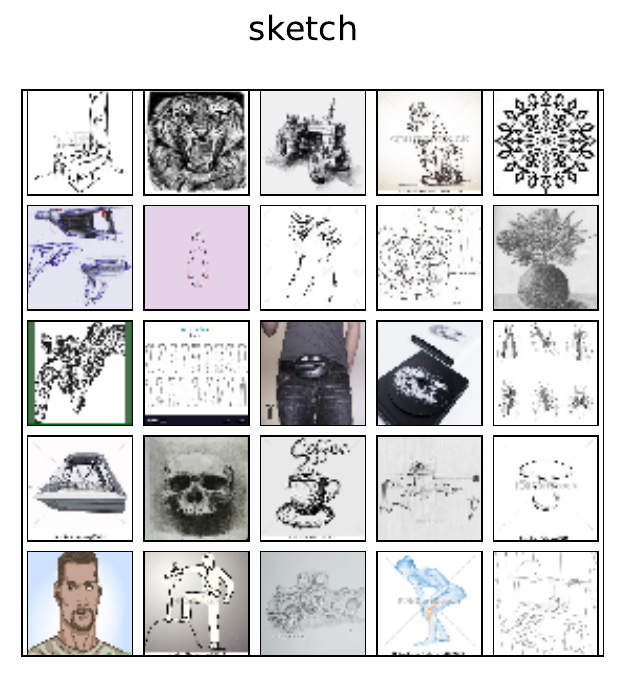}
        \caption[]%
        {{}}
        \label{fig:sketch}
    \end{subfigure}
    \begin{subfigure}[b]{0.19\textwidth}
        \centering
        \includegraphics[width=\textwidth]{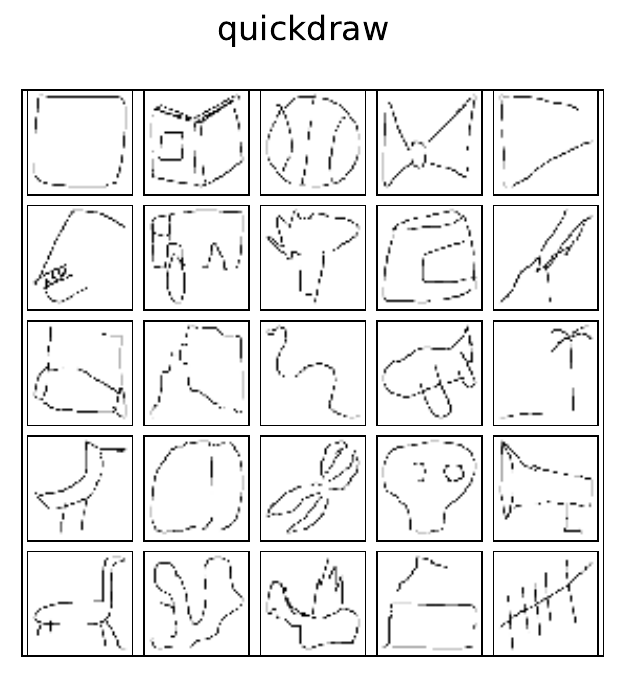}
        \caption[]%
        {{}}
        \label{fig:quickdraw}
    \end{subfigure}
    \caption[]
    {DomainNet~\cite{peng2019moment} qualitative examples}
    \label{fig:domainnet_qual}
\end{figure*}

\noindent \textbf{DomainNet.} For our primary experiments, we use the DomainNet~\cite{peng2019moment} dataset that consists of 0.6 million images spanning 6 domains, available at \href{http://ai.bu.edu/M3SDA/}{http://ai.bu.edu/M3SDA/}. For our experiments, we use 4 shifts from 5 domains: Real, Clipart, Sketch, Painting, and Quickdraw. Table~\ref{tab:domainnet_splits} summarizes the train/test statistics of each of these domains, while Fig.~\ref{fig:domainnet_qual} provides representative examples from each. As models use ImageNet initialization, we avoid using Real as a target domain.

\begin{table}[h]
    \centering
    \resizebox{\linewidth}{!}{
    \begin{tabular}{@{}lccccc@{}}
    \toprule
    & Real & Clipart & Painting  & Sketch & Quickdraw \\ \midrule
    Train & 120906 & 33525  & 48212 & 50416 & 120750 \\
    Test & 52041  & 14604 & 20916 & 21850 & 51750\\    
     \bottomrule
    \end{tabular}}
    \vspace{5pt}
    \caption{DomainNet~\cite{peng2019moment} train/test statistics}
    \label{tab:domainnet_splits}
\end{table}

\noindent \textbf{DIGITS.} We present results on the SVHN~\cite{netzer2011reading}$\rightarrow$MNIST~\cite{lecun1998gradient} domain shift. Both datasets consist of images of the digits 0-9. SVHN consists of 99289 (73257 train, 26032 test) RGB images whereas MNIST contains 70k (60k train, 10k test) grayscale images. Fig.~\ref{fig:digits_qual} shows representative examples. 

\subsection{Code and Implementation Details}
\label{sec:impl_details}

We use PyTorch~\cite{paszke2019pytorch} for all our experiments. Most experiments were run on an NVIDIA TitanX GPU. 

\par\noindent\textbf{CLUE}. We use the weighted K-Means implementation in scikit-learn~\cite{pedregosa2011scikit} to implement \texttt{CLUE}. Cluster centers are initialized via K-means++~\cite{david2007vassilvitskii}. The implementation uses the Elkan algorithm~\cite{hamerly2002alternatives} to solve K-Means. For $n$ objects, $k$ clusters, and $e$ iterations ($=300 $ in our experiments), the time complexity of the Elkan algorithm is roughly $\mathcal{O}(nke)$~\cite{elkan2003using}, while its space complexity is $\mathcal{O}(nk)$.

\par\noindent\textbf{DomainNet experiments}. We use a ResNet34~\cite{he2016deep} CNN architecture. For active DA (round 1 and onwards), we use the Adam~\cite{kingma2014adam} optimizer with a learning rate of 10$^{-5}$, weight decay of 10$^{-5}$ and train for 20 epochs per round (with an epoch defined as a complete pass over labeled target data) with a batch size of 64. For unsupervised adaptation (round 0), we use Adam with a learning rate of 3x10$^{-7}$, weight decay of 10$^{-5}$, and train for 50 epochs. Across all adaptation methods, we tune loss weights to ensure that the average labeled loss is approximately 10 times as large as the average unsupervised loss. We use random cropping and random horizontal flips for data augmentation. We set loss weights for supervised source training $\lambda_\source = 0.1$, supervised target training $\lambda_\mathcal{T} = 1$, and min-max entropy (for \texttt{MME}) $\lambda_{\mathcal{H}} = 0.1$. 

\begin{wrapfigure}{r}{0.45\linewidth}
    \begin{center}
    \includegraphics[width=0.9\linewidth]{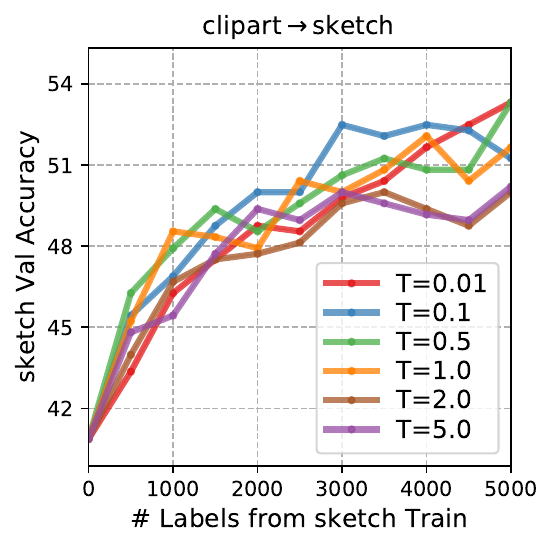}
    \caption{C$\rightarrow$S: Tuning softmax temperature with a small target validation set (1\% data).}%
        \vspace{-10pt}
    \label{fig:tune_temperature}
    \end{center}
\end{wrapfigure}
\par\noindent\textbf{Tuning softmax temperature.} In Active DA, it is unrealistic to assume access to a large validation set on the target to tune hyperparameters. To tune softmax temperature T for \method that trades off uncertainty and diversity, we thus create a small heldout validation set of just 1\% of target data (482 examples) on the Clipart$\rightarrow$Sketch shift, and perform a grid search over temperature values. We select T$=0.1$ based on its relatively consistent performance (Fig.~\ref{fig:tune_temperature}) across rounds on C$\to$S, and use it across shifts on DomainNet. We reiterate that T is an optional hyperparameter that may be tuned for a performance boost if a small validation set is available. As demonstrated in Sec 4.5 of the main paper, \method achieves SoTA results even with the default value of $T=1.0$. Further, we also find that T generalizes across shifts, suggesting that in practical scenarios it may be sufficient to tune it only on a single validation shift within a benchmark.

\noindent\textbf{DIGITS experiments.} We use the modified LeNet architecture proposed in Hoffman et al.~\cite{hoffman2017cycada} and exactly match the experimental setup in ~\texttt{AADA}~\cite{su2019active}. We use the Adam~\cite{kingma2014adam} optimizer with a learning rate of 2x10$^{-4}$, weight decay of 10$^{-5}$, batch size of 128, and perform 60 epochs of training per-round. We halve the learning rate every 20 epochs. We set loss weights  for supervised source training $\lambda_\source = 0.1$,  for supervised target training $\lambda_\mathcal{T} = 1$, and min-max entropy (for \texttt{MME}) $\lambda_{\mathcal{H}} = 1$. %

\begin{figure*}
    \centering
    \begin{subfigure}[b]{0.32\textwidth}  
        \centering 
        \includegraphics[width=\textwidth]{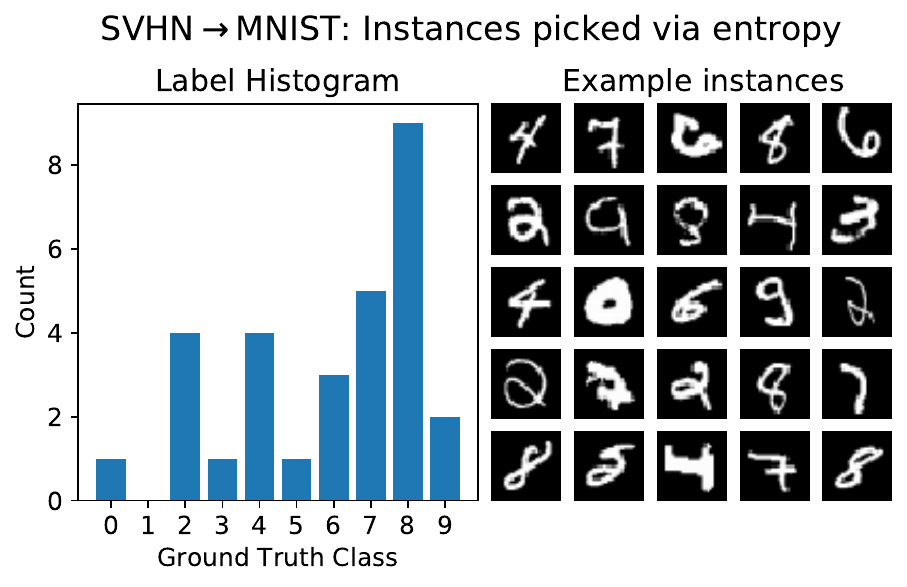}
        \caption[]%
        {{}}   
        \label{fig:ent_mme}
    \end{subfigure}
    \begin{subfigure}[b]{0.32\textwidth}
        \centering
        \includegraphics[width=\textwidth]{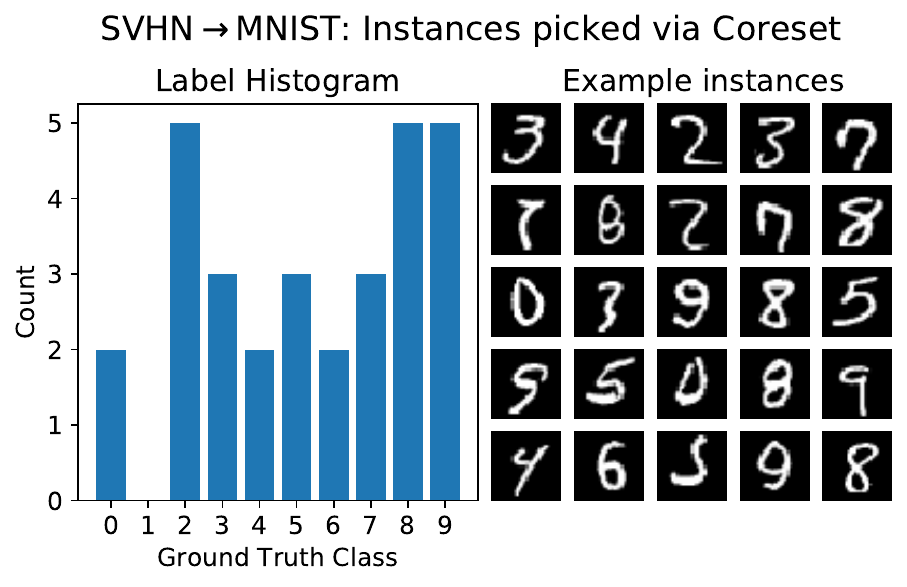}
        \caption[]%
        {{}}
        \label{fig:kcg_mme}
    \end{subfigure}
    \begin{subfigure}[b]{0.32\textwidth}
        \centering
        \includegraphics[width=\textwidth]{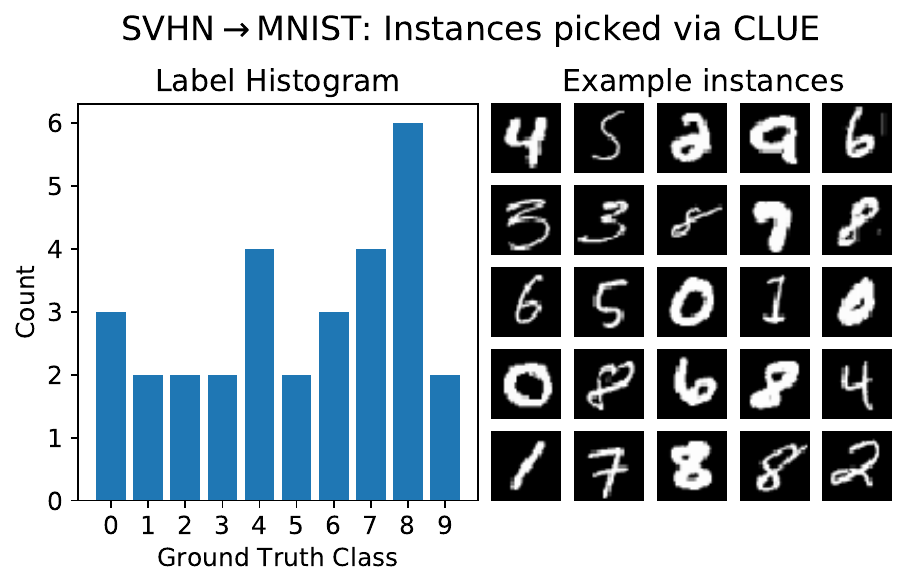}
        \caption[]%
        {{}}
        \label{fig:wks_mme}
    \end{subfigure}
    \caption[]
    {SVHN$\rightarrow$MNIST: Label histograms and examples of instances selected by entropy, coreset, and \method at Round 1 with $B=30$.}
    \label{fig:mme_picked}
\end{figure*}

\begin{figure*}
    \vspace{20pt}
        \centering
    \begin{subfigure}[b]{0.43\textwidth}   
        \centering 
        \includegraphics[width=\textwidth]{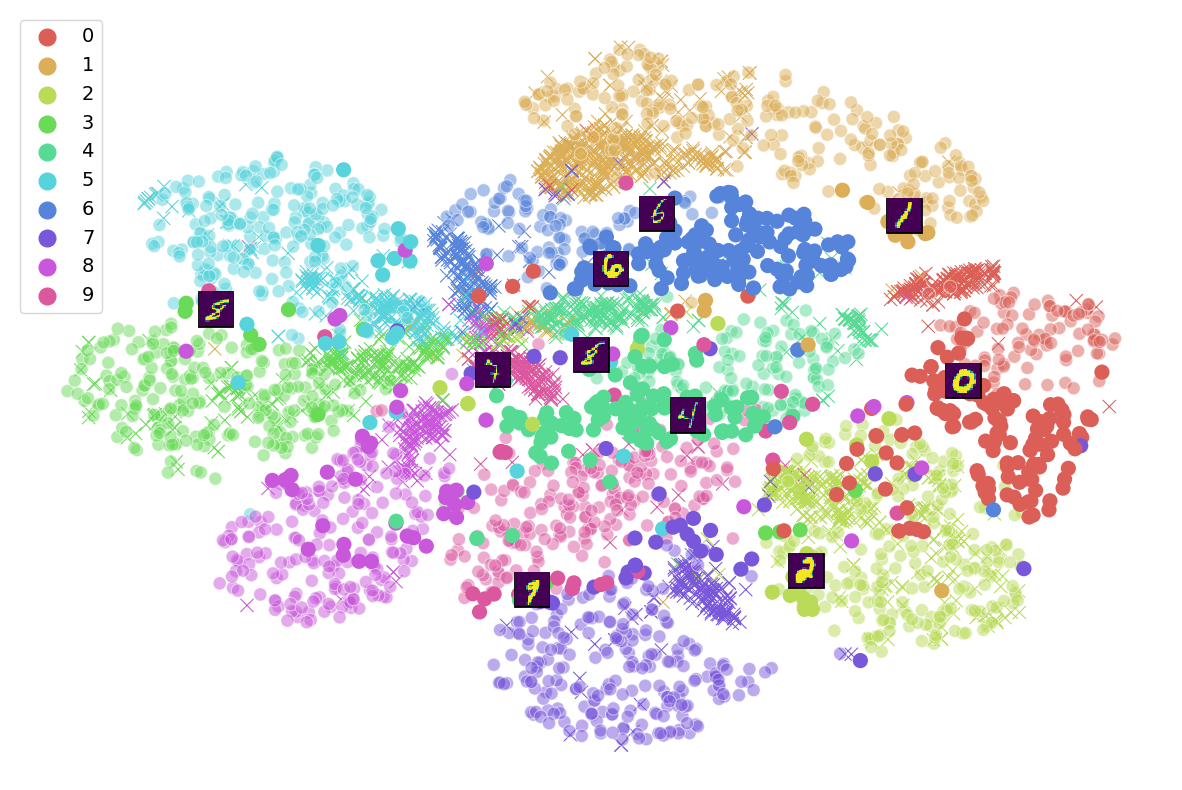}
        \caption[]%
        {{Round 1}}
        \label{fig:tsne_100}
    \end{subfigure}
    \begin{subfigure}[b]{0.43\textwidth}
     \centering 
     \includegraphics[width=\textwidth]{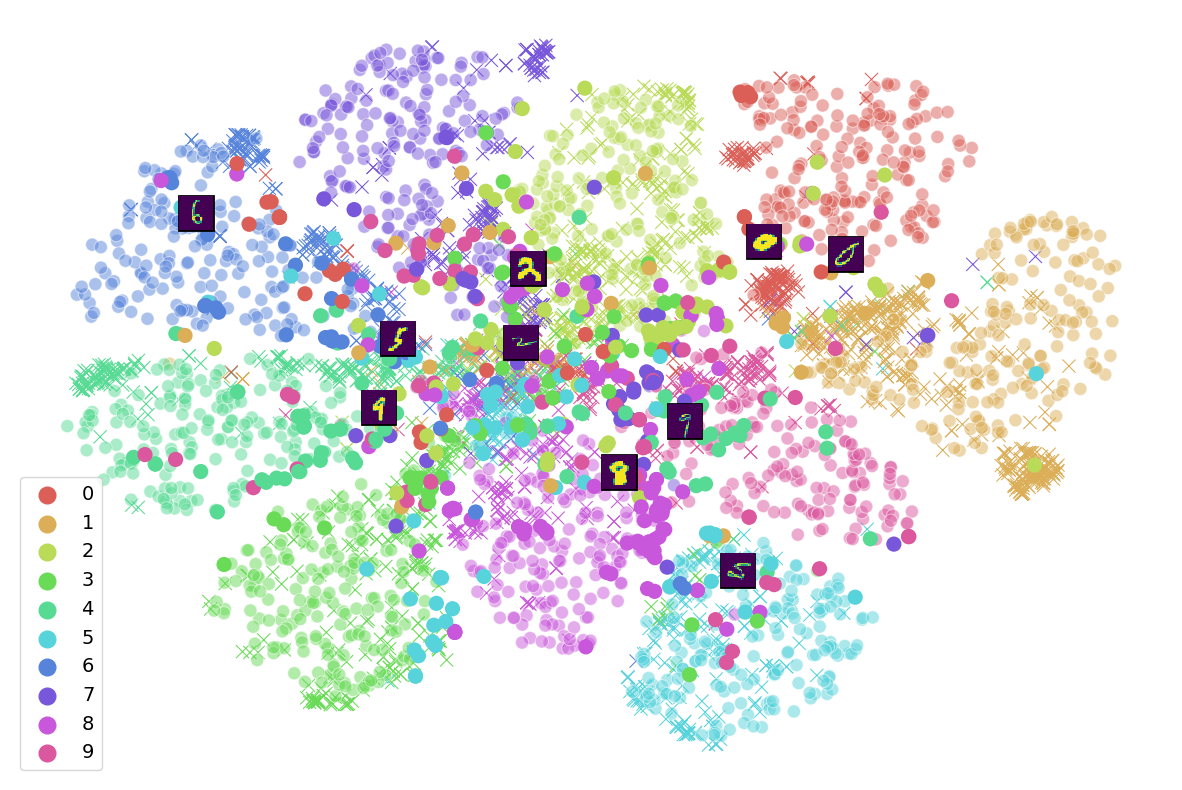}
     \caption[]
    {{Round 10}}
    \label{fig:tsne_200}
    \end{subfigure}
    \begin{subfigure}[b]{0.43\textwidth}
        \centering 
        \includegraphics[width=1\linewidth]{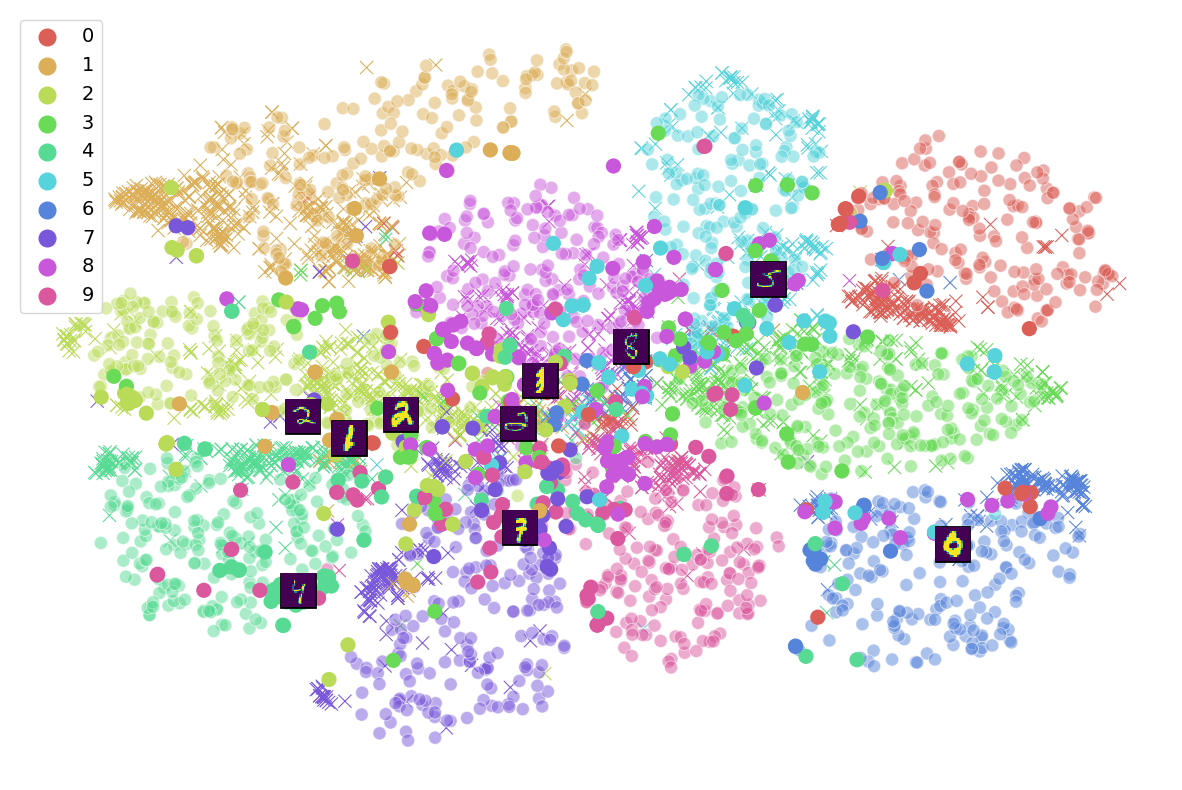}
        \caption[]
       {{Round 20}}
       \label{fig:tsne_200}
    \end{subfigure}    
    \begin{subfigure}[b]{0.43\textwidth}
        \centering 
        \includegraphics[width=1\linewidth]{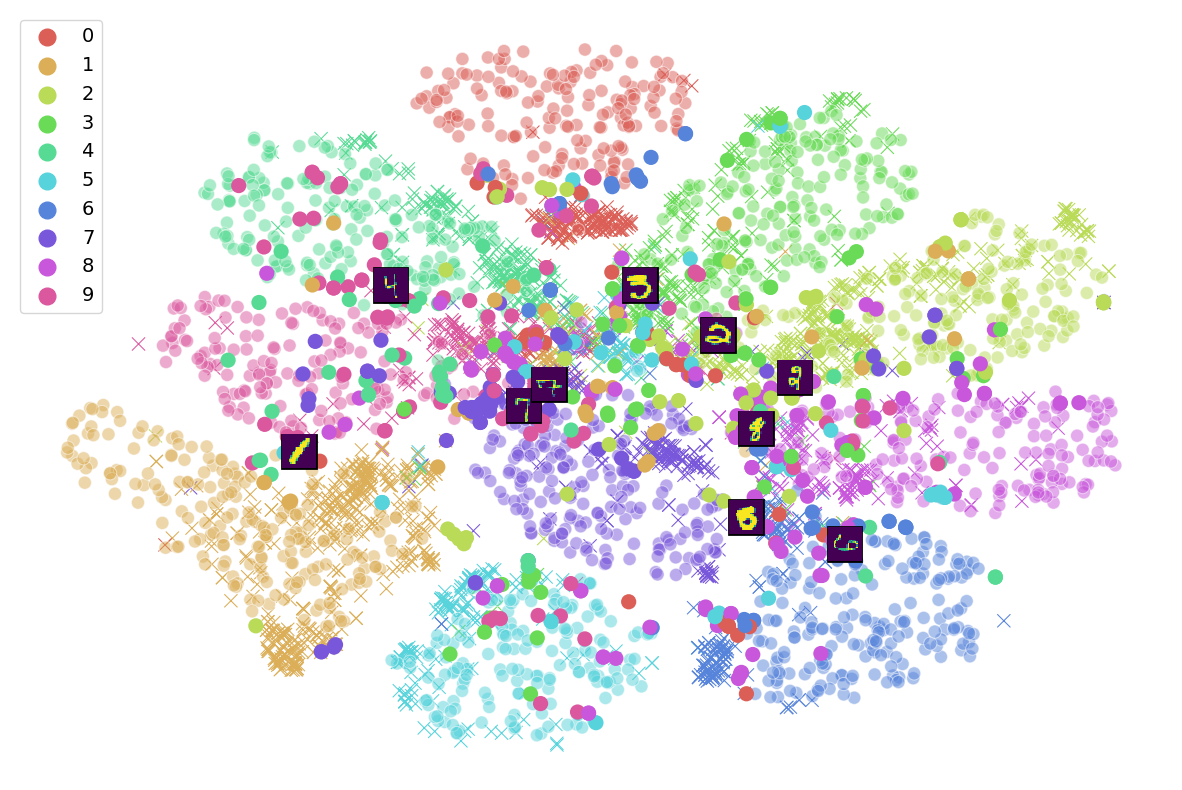}
        \caption[]
       {{Round 30}}
       \label{fig:tsne_300}
    \end{subfigure}    
    \caption[]{SVHN$\rightarrow$MNIST: TSNE visualization of feature space and instances picked by \method at rounds 1, 10, 20, and 30. Circles denote target points and crosses denote source points.} 
    \label{fig:tsne_digits}
\end{figure*}

\vspace{-10pt}
\subsubsection{Baseline Implementations}

We elaborate on our implementation of the \texttt{BADGE}~\cite{ash2019deep} and \texttt{AADA}~\cite{su2019active} baselines.

\noindent\texttt{BADGE.} BADGE ``gradient embeddings'' are computed by taking the gradient of model loss with respect to classifier weights, where the loss is computed as cross-entropy between the model's predictive distribution and its most confidently predicted class. Next, K-Means++~\cite{david2007vassilvitskii} is run on these embeddings to yield a batch of samples.

\noindent\texttt{AADA.} In \texttt{AADA}, a domain discriminator $G_d$ is learned to distinguish between source and target features obtained from an extractor $G_f$, in addition to a task classifier $G_y$. For active sampling, points are scored via the following importance weighting-based acquisition function ($\mathcal{H}$ denotes model entropy): $s(x)=\frac{1-G_{d}\left(G_{f}(x)\right)}{G_{d}^{*}\left(G_{f}(x)\right)} \mathcal{H}\left(G_{y}\left(G_{f}(x)\right)\right)$, and top $B$ instances are selected for labeling. In practice, to generate less redundant batches we randomly sample $B$ instances from the top-2\% scores, as recommended by the authors. Consistent with the original work, we also add an entropy minimization objective with a loss weight of 0.01.

\begin{figure}[h]
    \centering 
    \includegraphics[width=\linewidth]{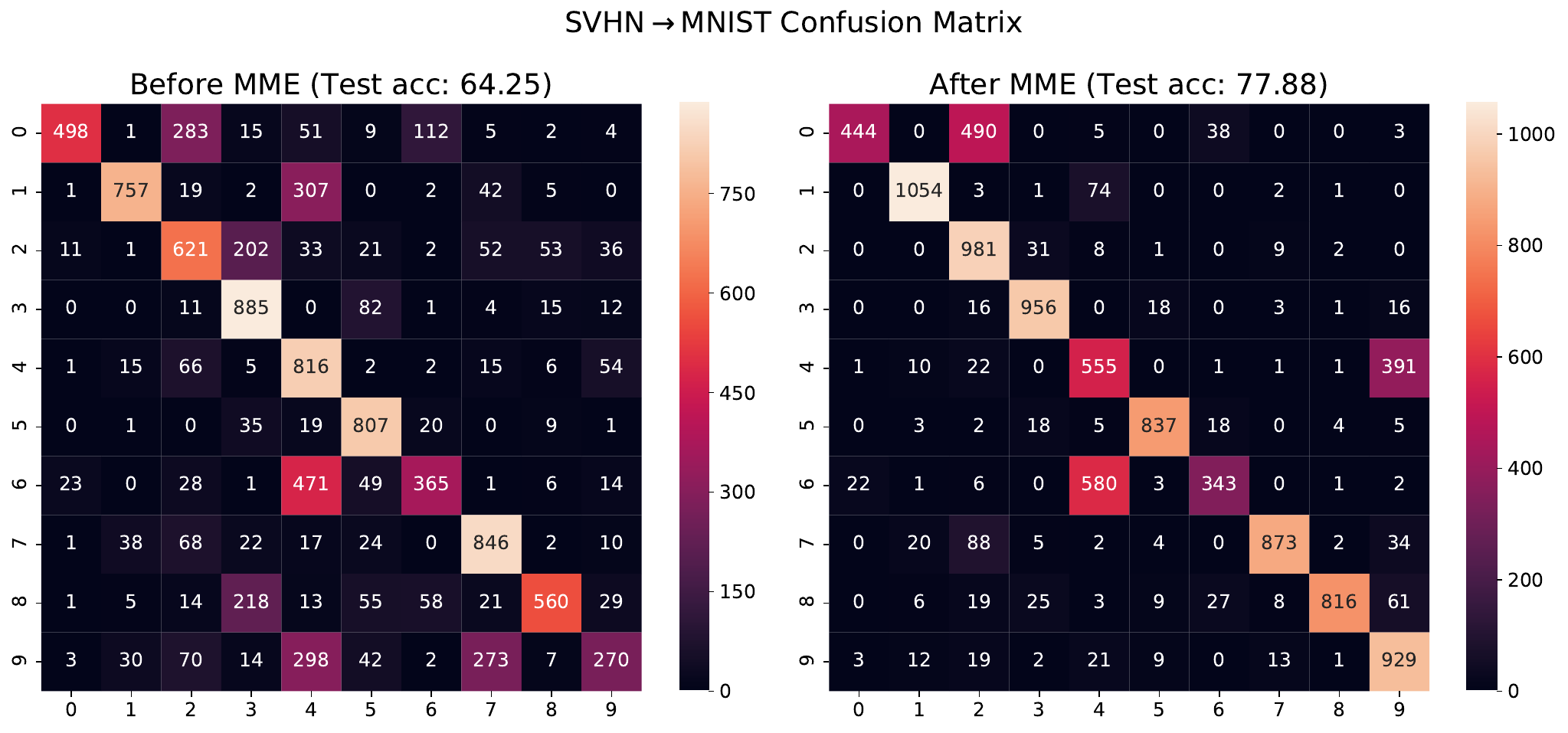}
    \caption{SVHN$\rightarrow$MNIST: Confusion matrix of model predictions before and after \texttt{MME} at round 0.}%
    \label{fig:results_mme_conf}
\end{figure}

\subsection{$\method$: Qualitative Analysis}
\label{sec:understand_adaclue}

In this section, we attempt to get a sense of the behavior of \method versus other methods via visualizations and qualitative examples on the SVHN$\rightarrow$MNIST shift. Fig.~\ref{fig:results_mme_conf} shows confusion matrices of model predictions before (\emph{left}) and after (\emph{right}) performing unsupervised adaptation (via \texttt{MME}) at round 0. As seen, \texttt{MME} aligns some classes (eg. 1's and 9's) remarkably well even without access to target labels. However, large misalignments remain for some other classes (0, 4, and 6).

\noindent\textbf{Visualizing selected points}. In Fig.~\ref{fig:mme_picked}, we visualize instances selected by three strategies at Round 0 -- entropy~\cite{wang2014new}, coreset~\cite{sener2017active}, and \method, with $B=30$. We visualize the ground truth label distribution of the selected instances, as well as qualitative examples. As seen, strategies vary across methods. ``Entropy'' tends to pick a large number of 8's, and selects high-entropy examples that (on average) appear challenging even to humans. ``Coreset'' tends to have a wider spread over classes. \method appears to interpolate between the behavior of these two methods, selecting a large number of 8's (like entropy) but also managing to sample atleast a few instances from every class (like coreset).

\noindent\textbf{t-SNE visualization over rounds.} In Fig.~\ref{fig:tsne_digits}, we illustrate the sampling behavior of \method over rounds via t-SNE~\cite{maaten2008visualizing} visualizations. We follow the same conventions as Fig.3 of the main paper, and visualize the logits of a subset of incorrect (large, opaque circles) and correct (partly transparent circles) model predictions on the target domain, along with instances sampled via \method. We oversample incorrect target predictions to emphasize regions of the feature space on which the model currently underperforms. Across all four stages, we find that \method samples instances that are uncertain (often present in a cluster of incorrectly classified instances) and diverse in feature space. This behavior is seen even at later rounds when classes appear better separated.

\subsection{Extended Description of the \method Objective}
\label{sec:wkm_proof}

We describe in more detail the \method objective presented (Eq. 4) in the main paper. Recall that we seek to identify target instances that are diverse in model feature space. Considering the L$_2$ distance in the CNN representation space $\featext(\cdot)$ as a dissimilarity measure, we quantify the dissimilarity between instances in a set $X_k$ in terms of its variance $\sigma^2(X_k)$ given by~\cite{zhang2012some}: 
\begin{equation}
    \label{eq:var}
    \begin{split}
\sigma^2(X_k) &= \frac{1}{2|X_k|^2}\sum_{\mathbf{x_i}, \mathbf{x_j} \in X_k}  || \phi(\mathbf{x_i}) - \phi(\mathbf{x_j})||^2 
\\
&= \frac{1}{|X_k|}\sum_{\mathbf{x} \in X_\cluster} ||\featext(\mathbf{x}) - \mathbf{\mu_k}||^2 
\\
\quad \mathrm{where} \quad 
\mathbf{\mu_k} &= \frac{1}{|X_k|}\sum_{\mathbf{x} \in X_\cluster} \featext(\mathbf{x})
\end{split}
\end{equation}
A small $\sigma^2(X_k)$ indicates that a set $X_k$ contains instances that are similar to one other. 
Our goal is to identify sets of instances that are representative of the unlabeled target set, by partitioning the unlabeled target data into $K$ sets, each with small $\sigma^2(X_k)$. 
Formulating this as a set-partitioning problem with partition function $\setpartition: X_T \rightarrow \{ X_1, X_2, ..., X_K \}$, we seek to find the $\setpartition$ that minimizes the sum of variance over all sets: 
\begin{equation}
    \argmin_\setpartition \sum_{\cluster=1}^{N} \sigma^2(X_k) 
    \label{eq:km}
\end{equation}
where $\sigma^2(X_k)$ is defined in Eq.~\ref{eq:var}. 

To ensure that the more informative/uncertain instances play a larger role in identifying representative instances, we employ weighted-variance, where an instance is weighted by its informativeness. Let $h_i$ denote the scalar weight corresponding to the instance $\mathbf{x_i}$. The weighted variance~\cite{price1972extension} $\sigma_\mathcal{H}^2(X_k)$ of a set of instances is given by: 
\begin{equation}
\begin{aligned}
& \sigma_\mathcal{H}^2(X_k) = \frac{1}{\sum_{\mathbf{x_i} \in X_\cluster} h_i}\sum_{\mathbf{x_i} \in X_\cluster} h_i||\featext(\mathbf{x_i}) - \mathbf{\mu_k}||^2 \\
& \quad \mathrm{where} \quad 
\mathbf{\mu_k} = \frac{1}{\sum h_i}\sum_{\mathbf{x_i} \in X_\cluster} h_i\featext(\mathbf{x_i})
\label{eq:wvar}
\end{aligned}
\end{equation}

Considering the informativeness (weight) of an instance to be its uncertainty under the model, given by $\mathcal{H}(Y|\mathbf{x})$ (defined in Eq. 1 in main paper), we rewrite the set-partitioning objective in Eq.~\ref{eq:km} to minimize sum of weighted variance of a set (from Eq.~\ref{eq:wvar}): 
\begin{equation}
    \label{eq:wkm}
    \begin{aligned}
    & \argmin_\setpartition \sum_{\cluster=1}^{K} \sigma_\mathcal{H}^2(X_k)  \\
    = & \argmin_\setpartition \sum_{\cluster=1}^{K} \frac{1}{Z_k}\sum_{\mathbf{x} \in X_\cluster} \mathcal{H}(Y|\mathbf{x}) ||\featext(\mathbf{x}) - \mathbf{\mu_k}||^2 \\
    &\mathrm{where} \;
    \mathbf{\mu_k} = \frac{1}{\sum_{x \in X_k} \mathcal{H}(Y|\mathbf{x})}\sum_{\mathbf{x} \in X_\cluster} \mathcal{H}(Y|\mathbf{x}) \featext(\mathbf{x}) \\
    &\mathrm{and} \quad
    Z_k = \sum_{x \in X_k} \mathcal{H}(Y|\mathbf{x})
    \end{aligned}
\end{equation}
This, gives us the overall set-partitioning objective for \method. %

\begin{figure*}
    \centering
    \begin{subfigure}[b]{0.9\textwidth}  
        \centering 
        \includegraphics[width=\textwidth]{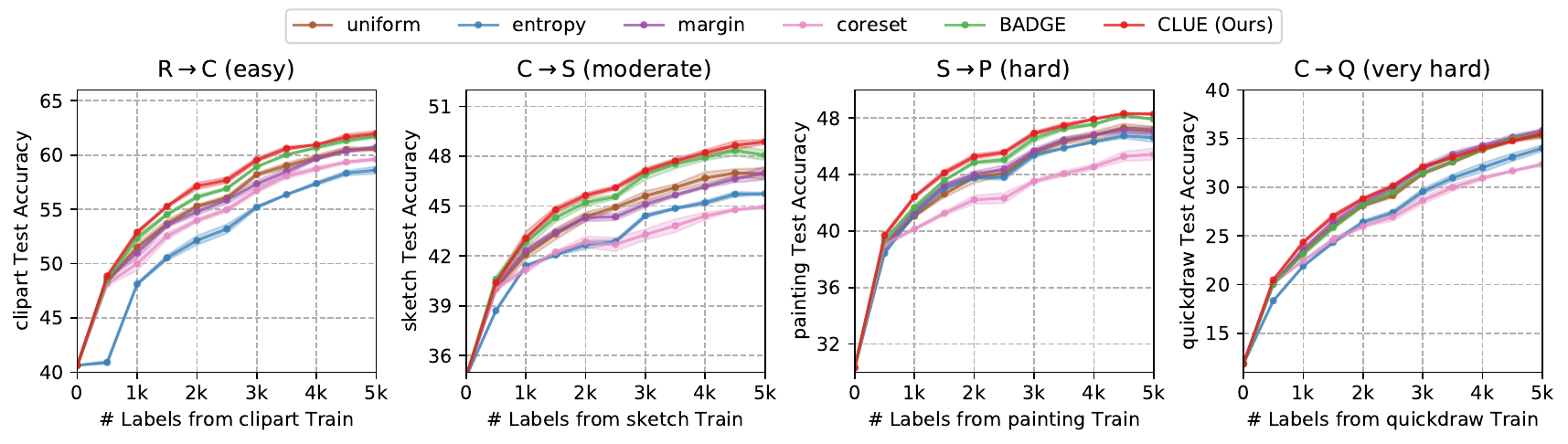}
        \caption[]%
        {{Finetuning (\texttt{ft}) a source model on target labels.}}   
        \label{fig:domainnet_ft}
    \end{subfigure}
    \begin{subfigure}[b]{0.9\textwidth}
        \centering
        \includegraphics[width=\textwidth]{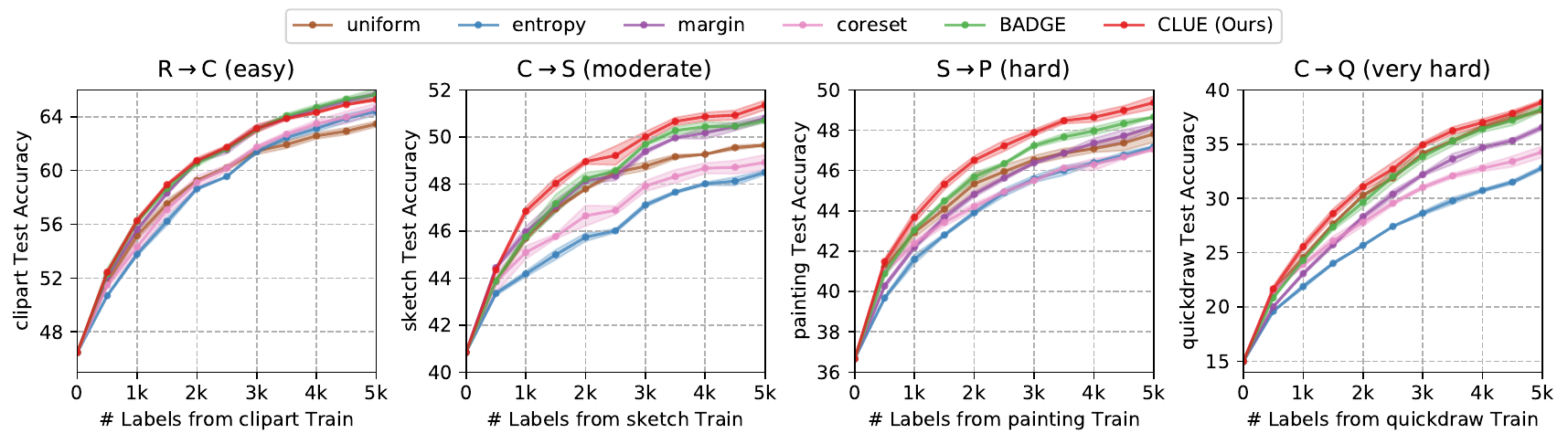}
        \caption[]%
        {{Semi-supervised DA via \texttt{MME}~\cite{saito2019semi} (state-of-the-art semi-supervised DA method), starting from a source model.}}
        \label{fig:domainnet_mme}
    \end{subfigure}
    \begin{subfigure}[b]{0.9\textwidth}
        \centering
        \includegraphics[width=\textwidth]{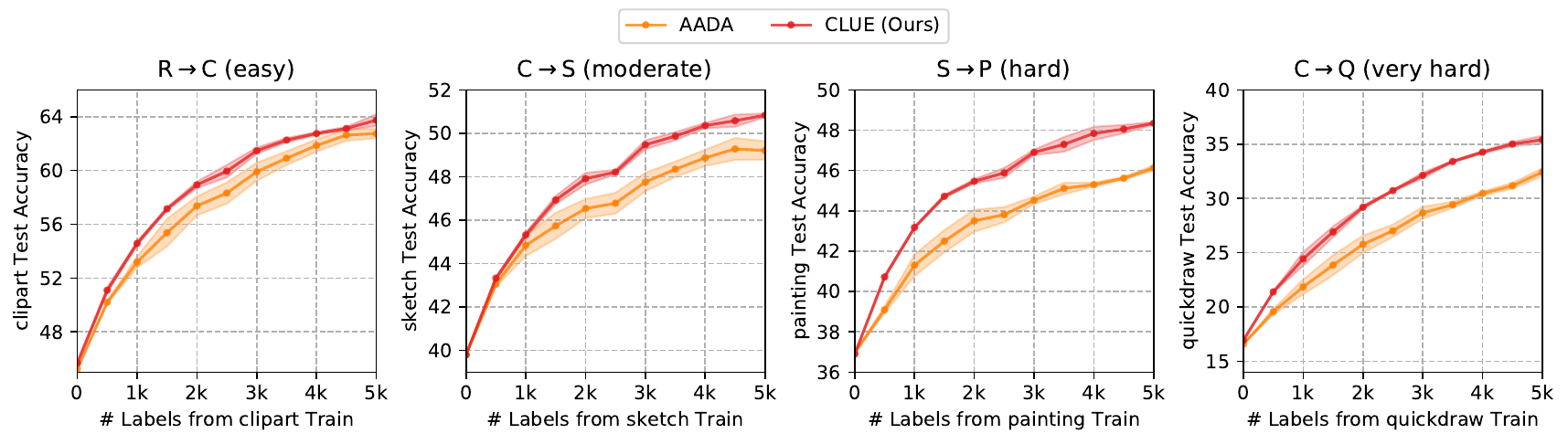}
        \caption[]%
        {{Semi-supervised DA via \texttt{DANN}~\cite{ganin2014unsupervised}, starting from a source model.}}
        \label{fig:domainnet_dann}
    \end{subfigure}
    \caption[]
    {Full plots for Active DA results on DomainNet, corresponding to Table 1 in the main paper. We plot accuracies on target test set for 4 DomainNet shifts of increasing difficulty spanning 5 domains: Real (R), Clipart (C), Sketch (S), Painting (P) and Quickdraw (Q). We perform 10 rounds of Active DA with $B=500$. We compare \method against state-of-the art methods for AL (\texttt{entropy}~\cite{wang2014new}, \texttt{margin}~\cite{roth2006margin}, \texttt{coreset}~\cite{sener2017active}, \texttt{BADGE}~\cite{ash2019deep}) and Active DA (\texttt{AADA}), spanning different AL paradigms: uncertainty sampling (U), diversity sampling (D), and hybrid (H) combinations of the two. We use multiple learning strategies: \textbf{(a)} finetuning (\texttt{ft}) from source, \textbf{(b)} \texttt{MME}~\cite{saito2019semi} (state-of-the-art semi-supervised DA method) from source, and \textbf{(c)} semi-supervised DA via \texttt{DANN}~\cite{ganin2014unsupervised} from source. Best viewed in color. We report accuracy mean and 1 standard deviation (via shading) over 3 runs.}
    \label{fig:domainnet}
\end{figure*}

\begin{figure*}
    \centering
    \begin{subfigure}[b]{0.25\textwidth}  
        \centering 
        \includegraphics[width=\textwidth]{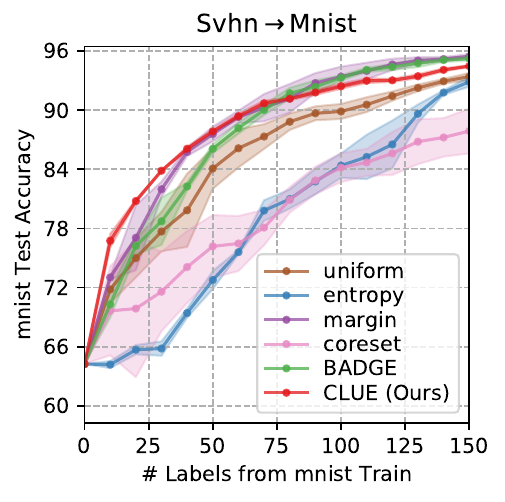}
        \caption[]%
        {{Finetuning (\texttt{ft}})}   
        \label{fig:s2m_ft}
    \end{subfigure}
    \begin{subfigure}[b]{0.25\textwidth}
        \centering
        \includegraphics[width=\textwidth]{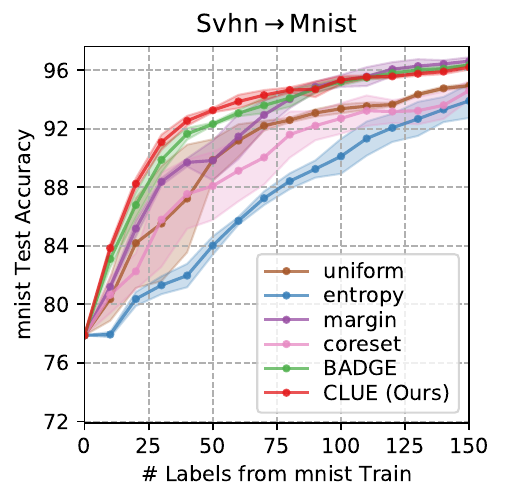}
        \caption[]%
        {{\texttt{MME}~\cite{saito2019semi}}}
        \label{fig:s2m_mme}
    \end{subfigure}
    \begin{subfigure}[b]{0.25\textwidth}
        \centering
        \includegraphics[width=\textwidth]{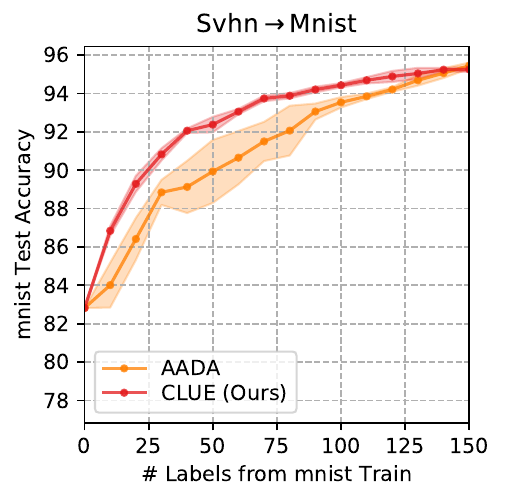}
        \caption[]%
        {{\texttt{DANN}~\cite{ganin2014unsupervised}}}
        \label{fig:s2m_dann}
    \end{subfigure}
    \caption[]
    {Full plots for Active DA results on SVHN$\to$MNIST (DIGITS) with $B=10$, corresponding to Table 2 (middle), in the main paper. We use multiple learning strategies: \textbf{(a)} finetuning (\texttt{ft}) from source, \textbf{(b)} \texttt{MME}~\cite{saito2019semi} (state-of-the-art semi-supervised DA method) from source, and \textbf{(c)} semi-supervised DA via \texttt{DANN}~\cite{ganin2014unsupervised} from source. Best viewed in color. We report accuracy mean and 1 standard deviation (via shading) over 3 runs.}
    \label{fig:s2m}
\end{figure*}

\begin{figure*}
    \centering
    \begin{subfigure}[b]{0.25\textwidth}  
        \centering 
        \includegraphics[width=\textwidth]{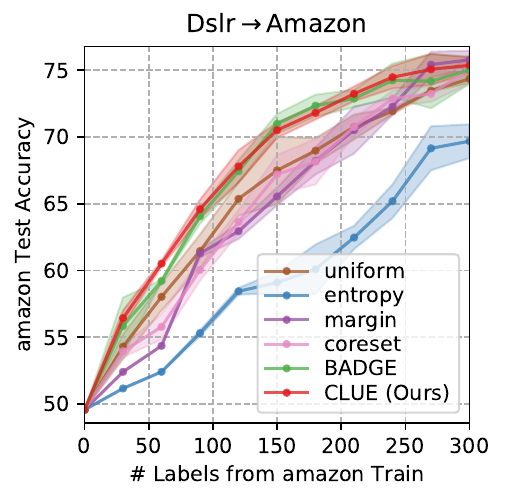}
        \caption[]%
        {{Finetuning (\texttt{ft})}}   
        \label{fig:d2a_ft}
    \end{subfigure}
    \begin{subfigure}[b]{0.25\textwidth}
        \centering
        \includegraphics[width=\textwidth]{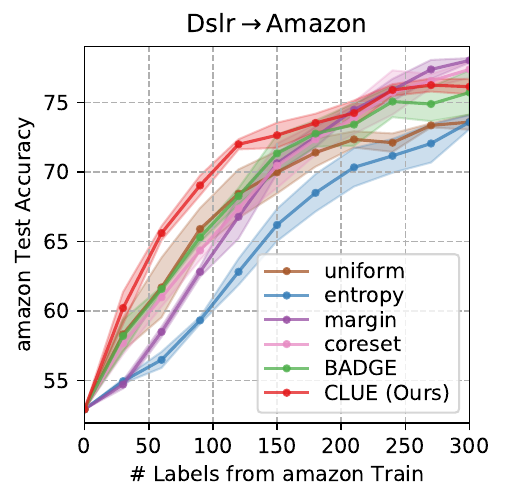}
        \caption[]%
        {{\texttt{MME}~\cite{saito2019semi}}}
        \label{fig:d2a_mme}
    \end{subfigure}
    \begin{subfigure}[b]{0.25\textwidth}
        \centering
        \includegraphics[width=\textwidth]{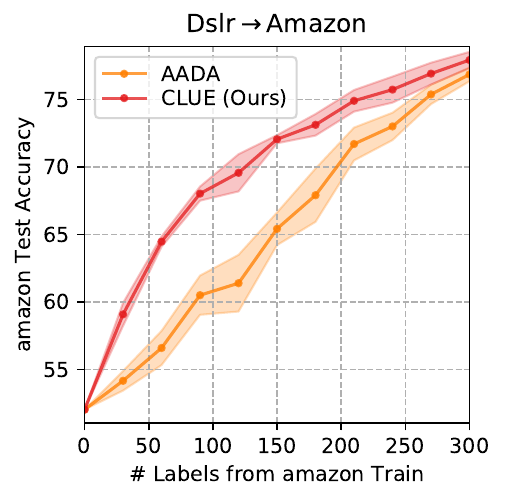}
        \caption[]%
        {{\texttt{DANN}~\cite{ganin2014unsupervised}}}
        \label{fig:d2a_dann}
    \end{subfigure}
    \caption[]
    {Full plots for Active DA results on DSLR$\to$Amazon (Office) with $B=30$, corresponding to Table 2 (right), in the main paper. We use multiple learning strategies: \textbf{(a)} finetuning (\texttt{ft}) from source, \textbf{(b)} \texttt{MME}~\cite{saito2019semi} (state-of-the-art semi-supervised DA method) from source, and \textbf{(c)} semi-supervised DA via \texttt{DANN}~\cite{ganin2014unsupervised} from source. Best viewed in color. At each round, we report accuracy mean and 1 standard deviation (via shading) over 3 runs.}
    \label{fig:d2a}
\end{figure*}

\subsection{Full performance plots}
\label{sec:al_ablation}

\noindent For ease of comparison, we presented performance at 3 intermediate sampling budgets in Tables 1 and 2 in the main paper. In Tables~\ref{fig:domainnet},~\ref{fig:d2a},~\ref{fig:s2m}, we include the full corresponding performance plots for completeness. Results are presented across 3 learning strategies: finetuning (\texttt{FT}) a source model, semi-supervised DA via \texttt{MME} starting from a source model, and semi-supervised DA via \texttt{DANN} starting from a source model.  As is common in active learning, we present results as learning curves, and report performance means and 1 standard deviation over 3 experimental runs via shading.

\subsection{Future Work}
\label{sec:future_work}

Our work suggests a few promising directions of future work. First, one could experiment with alternative uncertainty measures in \method instead of model entropy, including those (such as uncertainty from deep ensembles) that have been shown to be more reliable under a dataset shift~\cite{snoek2019can}. Further, one could incorporate specialized model architectures from few-shot learning~\cite{chen2019closerfewshot,saito2019semi} to deal with the label sparsity in the target domain. Finally, while we restrict our task to image classification in this paper, it is important to also study active domain adaptation in the context of related tasks such as object detection and semantic segmentation. 

\end{document}